\documentclass{article}
\usepackage{spconf,amsmath,graphicx}
\usepackage{multirow}
\usepackage{multicol}
\usepackage{url}
\usepackage{float}
\usepackage[]{subfigure}
\usepackage{epstopdf}

\usepackage[colorlinks,
pagebackref=true,
linkcolor=red,
anchorcolor=blue,
citecolor=green
]{hyperref}

\def \originalimage{I}
\def \sourceimage{{I}_{s}}
\def \targetimage{{I}_{t}}
\def \sourcegmimage{G_s}
\def \targetgmimage{G_t}
\def \line{\mathbf{l}}
\def \ppoint{\mathbf{p}}
\def \originvertex{\mathbf{V}}
\def \resultvertex{\hat{\mathbf{V}}}
\def \onevertex{\mathbf{v}}
\def \resultonevertex{\hat{\mathbf{v}}}
\def \bilinearweight{\mathbf{w}}

\title{Multiple Combined Constraints for Image Stitching}
\name{Kai Chen, Jingmin Tu, Binbin Xiang, Li Li, and Jian Yao$^{\dagger}$}
\address{School of Remote Sensing and Information Engineering, Wuhan University, P.R. China\\
$^{\dagger}$Email: \url{jian.yao@whu.edu.cn} Web: \url{http://cvrs.whu.edu.cn/}}
\begin{document}
\ninept
\maketitle
\begin{abstract}
	Several approaches to image stitching use different constraints to estimate the motion model between image pairs. These constraints can be roughly divided into two categories: geometric constraints and photometric constraints. In this paper, geometric and photometric constraints are combined to improve the alignment quality, which is based on the observation that these two kinds of constraints are complementary. On the one hand, geometric constraints (e.g., point and line correspondences) are usually spatially biased and are insufficient in some extreme scenes, while photometric constraints are always evenly and densely distributed. On the other hand, photometric constraints are sensitive to displacements and are not suitable for images with large parallaxes, while geometric constraints are usually imposed by feature matching and are more robust to handle parallaxes. The proposed method therefore combines them together in an efficient mesh-based image warping framework. It achieves better alignment quality than methods only with geometric constraints, and can handle larger parallax than photometric-constraint-based method. Experimental results on various images illustrate that the proposed method outperforms representative state-of-the-art image stitching methods reported in the literature.
\end{abstract}

\vspace{-5pt}
\section{Introduction}
\label{Sec:Introduction}
\vspace{-10pt}

Traditional image stitching methods estimate a global homography transformation to align image pairs~\cite{brown2007automatic}. The underlying two conditions (concentric condition and planar condition) are difficult to meet in practical applications. Recently spatially-varying motion model and mesh-based warping model are proposed to improve image stitching quality, especially for image pairs with large parallax. Compared with the global homography model, the higher degree of freedom makes them more flexible to handle large parallax, but also makes the model estimations more difficult. A lot of constraints are applied to restrain the process of model estimation, which can be roughly classified into two categories: geometric constraints and photometric constraints. Geometric constraints further include the point constraints and the line segment constraints.

A lot of methods use point features in the overlapping region as constraints to stitch image pairs~\cite{gao2011constructing,lin2011smoothly,zaragoza2013projective,chang2014shape,chai2016shape,zhang2014parallax}. Gao~\emph{et al.}~\cite{gao2011constructing} divided the image content into a ground plane and a distant plane. It estimated a two-homography model by point matches and align image pairs by fusing these two homographies according to a weight map. Lin~\emph{et al.}~\cite{lin2011smoothly} employed two sets of unmatched point features to jointly estimate the point correspondences and the spatially-varying affine transformation model simultaneously. Zaragoza~\emph{et al.}~\cite{zaragoza2013projective} extended the spatially-varying affine transformation to spatially-varying homography transformation. It handled stitching parallax by allowing local non-projective deviations apart from global homography transformation.

Point features are usually distributed with spatial bias~\cite{joo2015line} and are not sufficient enough to stably estimate the local warping model in some extreme cases (e.g., low-texture images). Therefore, a lot of methods~\cite{joo2015line,chen2016natural,hu2015multi,li2015dual,lin2016seagull,xiang2016locally,zhang2016multi} proposed to impose more constraints by line features. Joo~\emph{et al.}~\cite{joo2015line} proposed the line guided moving DLT (L-mDLT) method, which estimated a spatially-varying homography model with line correspondences. Similarly, Li~\emph{et al.}~\cite{li2015dual} proposed a mesh-based model by considering both point and line correspondences. Different from the above two methods who referred to the line constraints as data-term, In~\cite{lin2016seagull,chen2016natural}, line features are used as smoothness-term to preserve the line straightness before and after image warping.

Using the above two geometric constraints for image stitching is straightforward, but it strongly relies on abundant and robust results of feature extraction and matching, which are difficult to be ensured for images with complex structures. The photometric constraint is a good alternative that has been widely used in the problem of optical flow estimation. Recently, Lin~\emph{et al.}~\cite{lin2017direct} proposed to employ the photometric constraints to image stitching. It combined the advantage of dense photometric alignment with the efficiency of mesh-based image warping, and it obtained better alignment quality than geometric-constraint-based methods in both textured and low-textured images. However, the optimization process with photometric item was easily trapped into local minimum and the proposed method was not suitable for images with large parallax.

In this paper, we propose to augment the basic content-preserving warping (CPW)~\cite{liu2009content} framework by combining the photometric constraints with the geometric constraints. We observe that geometric and photometric constraints are actually complementary. On the one hand, geometric constraints are usually spatially biased and are insufficient in some extreme scenes, while photometric constraints are always evenly and densely distributed. On the other hand, photometric constraints are not suitable for images with large parallax, while geometric constraints imposed by feature matching are more robust to handle image parallax. We combine them in an unified CPW framework to achieve better alignment quality than geometric-constraint-based methods and handle larger parallax than photometric-constraint-based method. Abundant experiments demonstrate that our proposed method outperforms representative state-of-the-art image stitching methods.

\vspace{-5pt}
\section{Overview of The Proposed Scheme}
\label{Sec:GloballyConsistentAlignment}
\vspace{-10pt}
The proposed image stitching method adopts the two-stage pipeline which has been widely used in~\cite{hu2015multi,li2015dual,lin2017direct}. Firstly, a global parametric model is estimated from feature correspondences to roughly stitch image pairs. Next, the content-preserving warping (CPW)~\cite{liu2009content} is applied, which is a mesh-based model and can further align images in the overlapping region. Let $\originalimage_1$ and $\originalimage_2$ are a pair of images to be stitched, major steps of the proposed method include:\\
\\
\noindent (1) Estimate the global homography using point and line correspondences and apply global image warping. Specifically:\\
\begin{itemize}
	\setlength{\itemsep}{0pt}
	\setlength{\parsep}{0pt}
	\setlength{\parskip}{0pt}
	\vspace{-16pt}
	\item The point and line correspondences are firstly extracted from image pairs. In order to obtain point correspondences, we simply use SIFT~\cite{lowe2004distinctive} feature implemented by VLFeat~\cite{vedaldi2010vlfeat}. As for line segment, we adopt LSD~\cite{von2010lsd} to detect line segments and employ Line-point invariant (LPI)~\cite{fan2012robust} matching to obtain line correspondences.
	\item Secondly, we follow the process described in~\cite{li2015dual} to parameterize point and line features.
	\item Thirdly, the global homography $\mathbf{H}$ is estimated using direct linear transformation (DLT) with random sample consensus (RANSAC)~\cite{fischler1987random}.
	\item Lastly, $\originalimage_1$ and $\originalimage_2$ are globally stitched together.
\end{itemize}
The global alignment step would be beneficial to final stitching result in two aspects: For one thing, the global homography estimated from point and line features achieves better global alignment quality than the one estimated only from point features. For another, the process of RANSAC picks reliable matchings out from initial feature correspondences, which will be utilized in the next content-preserving warping.\\
\noindent (2) Apply CPW with multiple combined constraints over global aligned image pairs to locally align images. Let $\sourceimage$ and $\targetimage$ be the image pair after global alignment. The CPW achieves the local alignment by warping $\sourceimage$ to $\targetimage$. Similar to most existing methods, we construct a cost function considering the alignment quality as well as the smoothness of the mesh-based model and optimize it to obtain optimized positions of mesh vertexes.\\
\begin{itemize}
	\setlength{\itemsep}{0pt}
	\setlength{\parsep}{0pt}
	\setlength{\parskip}{0pt}
	\vspace{-12pt}
	\item Cost function. We propose to introduce geometric constraints and photometric constraints into CPW framework simultaneously. Overall, three data terms and two smoothness terms are considered in the cost function, which are point correspondence term (Section~\ref{Subsec:PCT}), line correspondence term (Section~\ref{Subsec:LCT}), photometric term (Section~\ref{Subsec:PT}), similarity transformation term (Section~\ref{Subsec:ST}) and line collinearity term (Section~\ref{Subsec:LCLT}).
	\item Optimization. As the photometric constraints are related to image content, it will get updated after each CPW. Therefore, we perform the process of CPW multiple times until the mesh becomes stable. In order to stitch images with large parallax, similar to~\cite{lin2017direct}, we also adopt a coarse-to-fine optimization scheme on Gaussian pyramid images.
\end{itemize}
After the optimization, local homography can be inferred from the transformation of each mesh vertex, based on which $\sourceimage$ are warped to $\hat{\sourceimage}$ to be locally aligned to $\targetimage$.\\
\noindent (3)  Blend the target image $\targetimage$ and the warped source image $\hat{\sourceimage}$ linearly to obtain the resultant stitching image. In this paper, we simply use linear blending to illustrate the misalignments and make comparisons with other algorithms. Actually, many advanced blending method (e.g., multi-band blending~\cite{burt1983multiresolution}) can be employed to get better stitching results.

We will describe the proposed CPW with multiple combined constraints in more detail in Section~\ref{Sec:CPW}.

\vspace{-5pt}
\section{Multiple Constraints for Content Preserving Warping}
\label{Sec:CPW}
\vspace{-5pt}

Images after global alignment have been roughly stitched together. In the following content-preserving warping (CPW) step, we first divide $\sourceimage$ into a $m\times n$ regular mesh ($m=n=32$ in our implementation), where the original coordinates of mesh vertexes are denoted as $\originvertex$. The objective of CPW is to obtain the optimized positions of mesh vertexes $\resultvertex$.

In order to accomplish this target and estimate the mesh-based model robustly, we propose to combine the geometric constraints with the photometric constraints in the cost function of CPW. Specifically, point correspondence, line correspondences and photometric terms are three data terms to ensure that the optimized mesh can eliminate the misalignment in the overlapping region. Similarity transformation and line collinearity terms are two smoothness terms that are used for shape preservation.
\vspace{-8pt}
\subsection{Point Correspondence Term}
\label{Subsec:PCT}
\vspace{-3pt}

We restrict the matching feature points between $\sourceimage$ and $\targetimage$ to be warped to close positions to improve local alignment quality in image overlapping region. Let $\ppoint$ and $\ppoint^{'}$ be a pair of matching points on $\sourceimage$ and $\targetimage$ respectively. We first denote $\ppoint$ with four vertexes of the quad containing $\ppoint$ by bilinear interpolation. Let $\originvertex_{\ppoint}=[\onevertex_{\ppoint}^1,\onevertex_{\ppoint}^2,\onevertex_{\ppoint}^3,\onevertex_{\ppoint}^4]$ be the corresponding four vertexes, and $\bilinearweight_{\ppoint}=[\omega_{\ppoint}^1,\omega_{\ppoint}^2,\omega_{\ppoint}^3,\omega_{\ppoint}^4]^{\mathsf{T}}$ be their bilinear interpolation coefficients. $\ppoint$ thus can be expressed as $\ppoint=\originvertex_{\ppoint}\bilinearweight_{\ppoint}$. We assume that the interpolation coefficients $\bilinearweight_{\ppoint}$ is consistent before and after CPW. We therefore construct the point correspondence term by summing alignment errors over all matching feature points.
\vspace{-5pt}
\begin{equation}\label{Eq:PCT}
	E_{f}(\resultvertex)=\begin{matrix}
	\sum_{i}\|\resultvertex_{\ppoint_i}\bilinearweight_{\ppoint_i}-\ppoint_{i}^{'} \|^2
	\end{matrix},\vspace{-5pt}
\end{equation}
where $\ppoint_i$ and $\ppoint_i^{'}$ denote the $i$-th pair of matching points.
\vspace{-5pt}
\subsection{Line Correspondence Term}
\label{Subsec:LCT}
\vspace{-3pt}

Similar to the point correspondence term, we restrict the matching lines to be close after local warping. Suppose that $\line=[a,b,c]$ and $\line^{'}=[a^{'},b^{'},c^{'}]$ are a pair of matching lines of $\sourceimage$ and $\targetimage$. In order to measure the alignment error between $\line$ and $\line^{'}$, we uniformly sample key points along $\line$ and require the distance from all key points to $\line^{'}$ to be minimized. Therefore, the line correspondence term is defined as:
\vspace{-5pt}
\begin{equation}\label{Eq:LCT}
	E_{l}(\resultvertex)=\begin{matrix}
	\sum_{j,k}\|({\line_{j}^{'}}^{\mathsf{T}}\cdot \resultvertex{\ppoint_{j,k}}\bilinearweight_{\ppoint_{j,k}})/\sqrt{{a_j^{'}}^2+{b_j^{'}}^2} \|^2
	\end{matrix},\vspace{-5pt}
\end{equation}
where $\ppoint_{j,k}$ denotes the $k$-th key point of $j$-th pair of matching lines. $\resultvertex_{\ppoint_{j,k}}$ are four mesh vertexes of the quad enclosing $\ppoint_{j,k}$ and $\bilinearweight_{\ppoint_{j,k}}$ are the corresponding bilinear interpolation coefficients.

\vspace{-5pt}
\subsection{Photometric Term}
\label{Subsec:PT}
\vspace{-3pt}
Above two geometric constraints are widely used to mesh-based model estimation. Recently, Lin~\emph{et al.}~\cite{lin2017direct} adopt a photometric constraint to align images. Their proposed photometric constraint is defined as:
\vspace{-5pt}
\begin{equation}\label{Eq:PTM}
E_{pm}=\begin{matrix}
\sum_{i}\|\targetimage(\ppoint_i+\tau(\ppoint_i))- \sourceimage(\ppoint_i)\|^2
\end{matrix},\vspace{-5pt}
\end{equation}
where $\ppoint_i$ is a set of uniformly sampled points in overlapping region. $\sourceimage(\cdot)$ and $\targetimage(\cdot)$ denote image intensity of source image and target image respectively. They parameterize the offset $\tau(\ppoint_i)$ of sampled point $\ppoint_i$ using offsets of four mesh vertexes enclosing $\ppoint_i$ and relate the minimization of Eq.~\ref{Eq:PTM} to mesh-based optimization. We append a gradient-based component to Eq.~\ref{Eq:PTM} to beyond the illumination constancy to form our photometric constraint term:
\vspace{-5pt}
\begin{equation}\label{Eq:PT}
E_{p}(\resultvertex)=E_{pm}+\lambda E_{pg},\vspace{-5pt}
\end{equation}
where $E_{pg}$ is the gradient component of our photometric constraint and $\lambda$ is the weight to balance intensity and gradient components ($\lambda=1$ in our implementation). $E_{pg}$ is defined as:
\vspace{-5pt}
\begin{equation}\label{Eq:PTG}
E_{pg}=\begin{matrix}
\sum_{i}\|\targetgmimage(\ppoint_i+\tau(\ppoint_i))-\sourcegmimage(\ppoint_i) \|^2
\end{matrix}.\vspace{-5pt}
\end{equation}

It is to be noted that, in order to achieve rotational invariance~\cite{fortun2015optical}, we set $G_{s,t}=\|\nabla I_{s,t} \|$ rather than $G_{s,t}=\nabla I_{s,t}$.

\vspace{-5pt}
\subsection{Similarity Transformation Term}
\label{Subsec:ST}
\vspace{-3pt}

We define our first smoothness term similar to the one of~\cite{liu2009content}, which constrains the similarity transformation for each quad. For each quad $[\onevertex_1,\onevertex_2,\onevertex_3,\onevertex_4]$ in current mesh grid, it can be divided into two triangulations $\triangle\onevertex_1\onevertex_3\onevertex_2$ and $\triangle\onevertex_4\onevertex_3\onevertex_2$. Vertex $\onevertex_1$ of $\triangle\onevertex_1\onevertex_3\onevertex_2$ can be represented as:
\vspace{-9pt}
\begin{equation}\label{Eq:ST}
	\onevertex_1=\onevertex_2+u(\onevertex_3-\onevertex_2)+v\mathbf{R}_{90}(\onevertex_3-\onevertex_2),\mathbf{R}_{90}=\begin{bmatrix}
	0 & 1 \\ -1 & 0
	\end{bmatrix},\vspace{-9pt}
\end{equation}
where $u$ and $v$ can be computed from original positions of $\onevertex_1$, $\onevertex_2$ and $\onevertex_3$. To constrain each quad to undergo a similarity transformation, for the warped triangulation $\triangle\resultonevertex_1\resultonevertex_3\resultonevertex_2$, $\resultonevertex_1$ should be ensured to be represented by $\resultonevertex_2$ and $\resultonevertex_3$ using the same local coordinates $(u,v)$ that computed from original positions. Therefore, the overall similarity transformation term is defined as:
\vspace{-5pt}
\begin{equation}\label{Eq:STT}
	E_{s}(\resultvertex)=\begin{matrix}
	\sum_{i}^{N_t}\|\resultonevertex_1^{i}-(\resultonevertex_2^i+u(\resultonevertex_3^i-\resultonevertex_2^i)+v\mathbf{R}_{90}(\resultonevertex_3^i-\resultonevertex_2^i)) \|^2
	\end{matrix},\vspace{-5pt}
\end{equation}
where $N_t$ is the total number of triangulations in the mesh grid.

\vspace{-5pt}
\subsection{Line Collinearity Term}
\label{Subsec:LCLT}
\vspace{-3pt}

We construct another smoothness term based the extracted line segments. Unlike the cost function proposed in~\cite{li2015dual}, in which only matched lines are utilized to constrain the cost function, we use the matched lines as one of our data term (Section~\ref{Subsec:LCT}) and use all extracted lines (matched and unmatched) to define our line collinearity term. Similar to Section~\ref{Subsec:LCT}, for each line segment $\line$ on $\sourceimage$, its two endpoints are denoted as $\ppoint_e$ and $\ppoint_s$. We uniformly sample key points along $\line$ and compute its $1D$ coordinate $u$ in the local coordinate system defined by $\ppoint_e$ and $\ppoint_s$. To preserve the straightness of line segment, each key point is supposed to be represented by the same local coordinate after mesh-based warping. Therefore, the line collinearity constraint term is defined as:
\vspace{-5pt}
\begin{equation}\label{Eq:LCLT}
E_{c}(\resultvertex)=\begin{matrix}
\sum_{i}^{N_l}\sum_{j}^{N_p^i}\|\ppoint_j^i-(\ppoint_s^i+u(\ppoint_e^i-\ppoint_s^i)) \|^2
\end{matrix},\vspace{-5pt}
\end{equation}
where $\ppoint_e^i$ and $\ppoint_s^i$ are two endpoints of $i$-th line segment. $\ppoint_j^i$ is the $j$-th key point of the $i$-th line segment. $N_p^i$ denotes the number of key points that the $i$-th line contains and $N_l$ is the total number of extracted line segments of $\sourceimage$. The key point and endpoints $\ppoint_e^i$ and $\ppoint_s^i$ are further parameterized by the mesh vertexes using bilinear interpolation, which is similar to the process described in previous section.
\vspace{-5pt}
\subsection{Overall Cost Function}
\label{Subsec:COST}
\vspace{-3pt}
We append a weight factor to each constraint term to form our final cost function:
\vspace{-5pt}
\begin{equation}\label{Eq:TOTAL}
E=\alpha E_f + \beta E_l + \gamma E_p + \delta E_c + \eta E_s.\vspace{-5pt}
\end{equation}
In Eq.~\ref{Eq:TOTAL}, $\eta$ is set to $0.2$ and another four weights are all set to $1.0$ in our implementation. The overall cost function is quadratic and can be easily minimized by any sparse linear system. We minimize the cost function defined by Eq.~\ref{Eq:TOTAL} multiple times and after each minimization, we update the photometric term $E_p$ and continue to the next minimization. We consider the optimization to have converged if the average change of vertex positions between adjacent two optimizations is smaller than one pixel. Finally, we use the optimized mesh to warp $\sourceimage$ to $\hat{\sourceimage}$ locally and blend it with $\targetimage$ to get the stitching result.

\vspace{-10pt}
\section{Experimental Results}
\label{Sec:ExperimentalResults}
\vspace{-10pt}

In order to demonstrate the superiority of our proposed image stitching method, we quantitatively compare it with three state-of-the-are methods: point-correspondence-based APAP~\cite{zaragoza2013projective}, dual-feature-based DF-W~\cite{li2015dual} and photometric-based MPA~\cite{lin2017direct}. After that, we conduct another experiment to qualitatively compare the stitching results produced by our method with different constraint combinations. 

\vspace{-5pt}
\subsection{Quantitative Comparison}
\label{Subsec:Quantitative}
\vspace{-5pt}

To quantitatively measure the alignment accuracy of image stitching, we adopt the same measurement in~\cite{li2015dual,lin2017direct} to compute the stitching quality of each pair of images. Specifically, we compute the RMSE of one minus normalized cross correlation (NCC) over a local $w\times w$ window for all pixel in the overlapping region to obtain the stitching quality metric:
\vspace{-10pt}
\begin{equation}\label{Eq:RMSE}
RMSE(\hat{\sourceimage},\targetimage)=\sqrt{\frac{1}{N}\begin{matrix}
	\sum_{\Omega}(1.0-NCC(\ppoint_s,\ppoint_t))
	\end{matrix}},\vspace{-8pt}
\end{equation}
where $\Omega$ denotes the overlapping region of $\hat{\sourceimage}$ and $\targetimage$. $N$ is the number of pixels in $\Omega$. $\ppoint_s$ and $\ppoint_t$ are the corresponding pixels in $\hat{\sourceimage}$ and $\targetimage$ respectively.

\begin{figure}[t]
	\centering
	\begin{minipage}[t]{1.0\columnwidth}
		\begin{minipage}[t]{0.24\columnwidth}
			\includegraphics[width=1.0\textwidth]{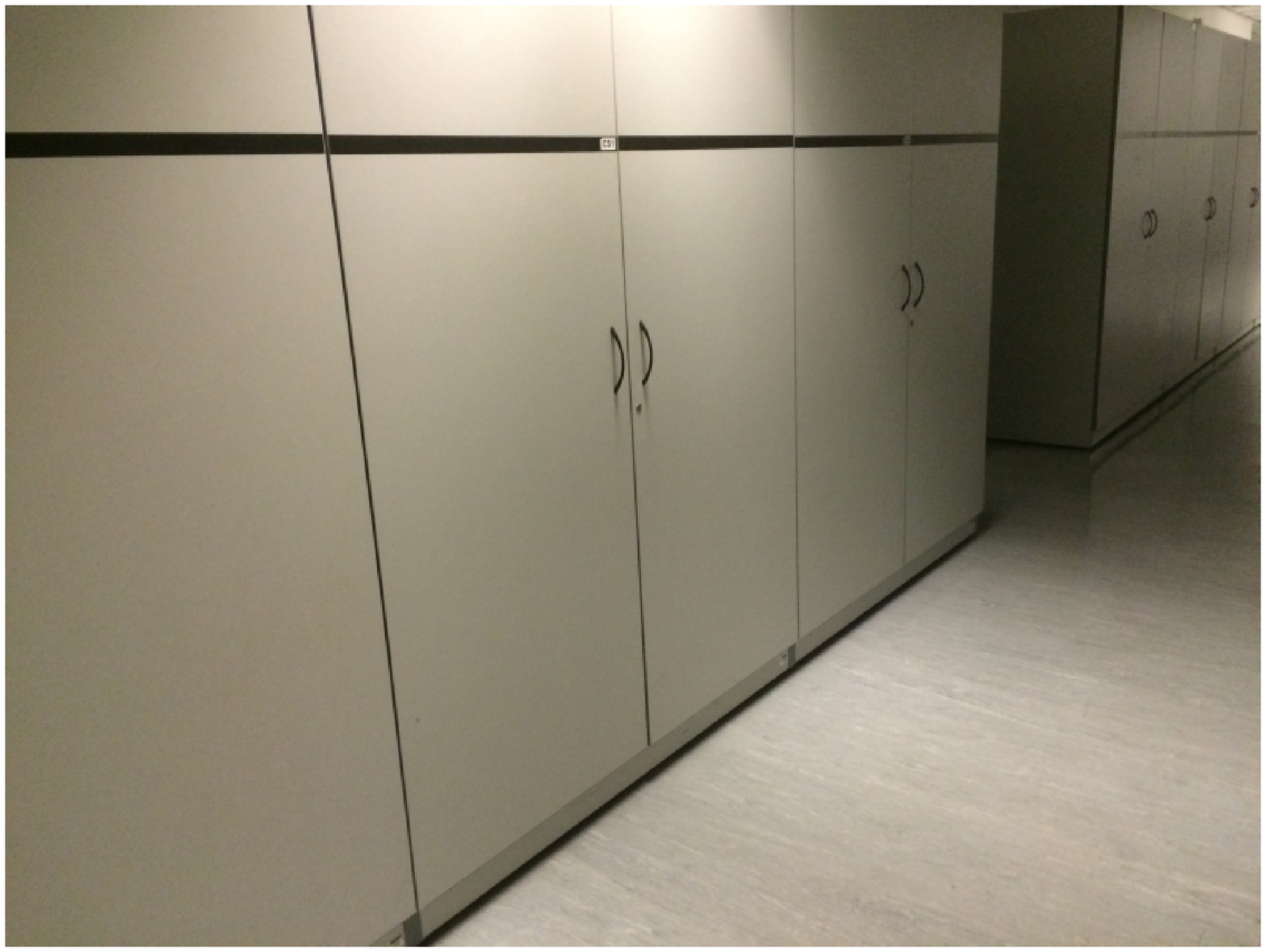}
			\centering{\emph{cabinet}}
		\end{minipage}
		\hfill
		\begin{minipage}[t]{0.24\columnwidth}
			\includegraphics[width=1.0\textwidth]{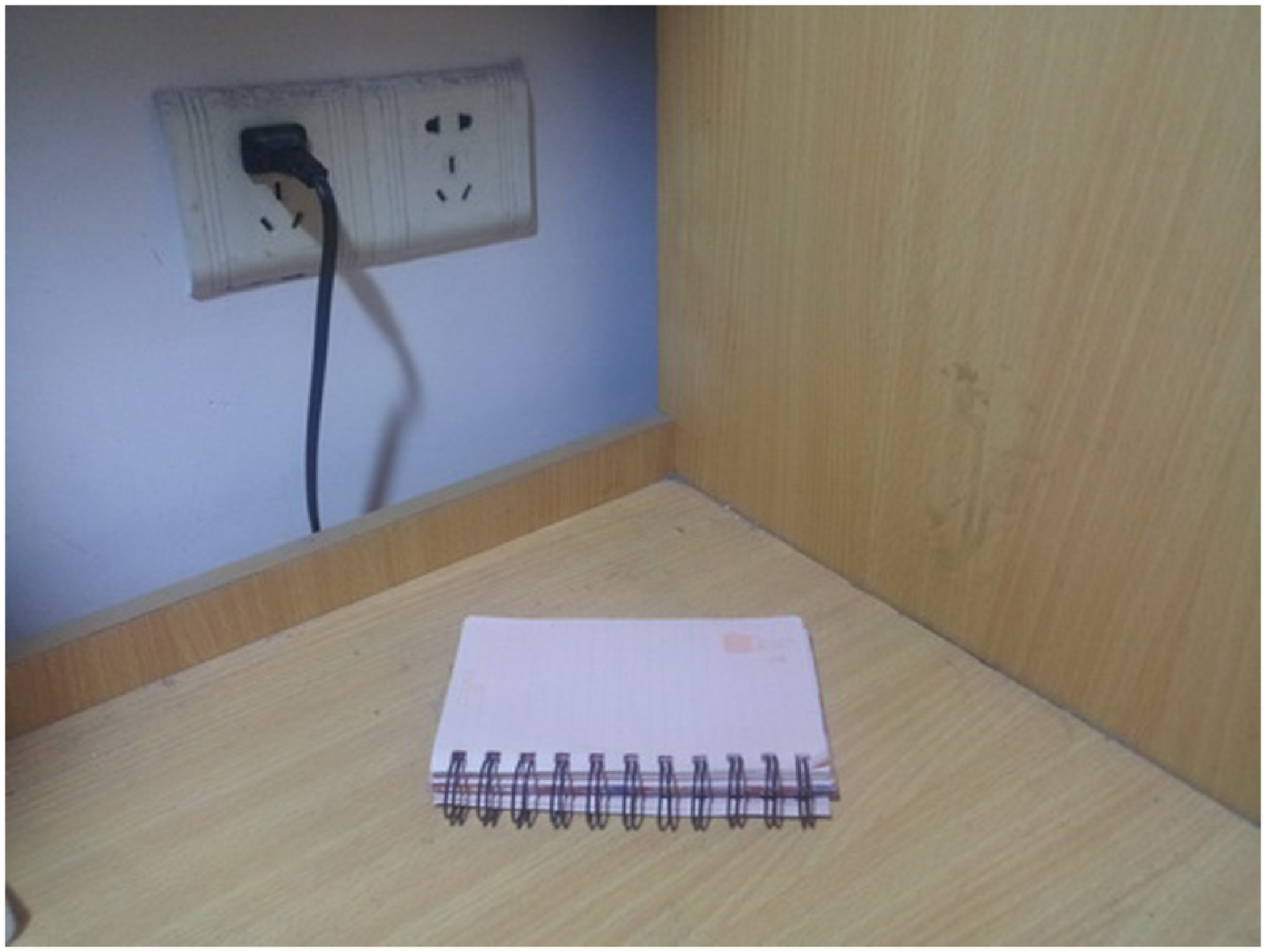}
			\centering{\emph{desk}}
		\end{minipage}
		\hfill
		\begin{minipage}[t]{0.24\columnwidth}
			\includegraphics[width=1.0\textwidth]{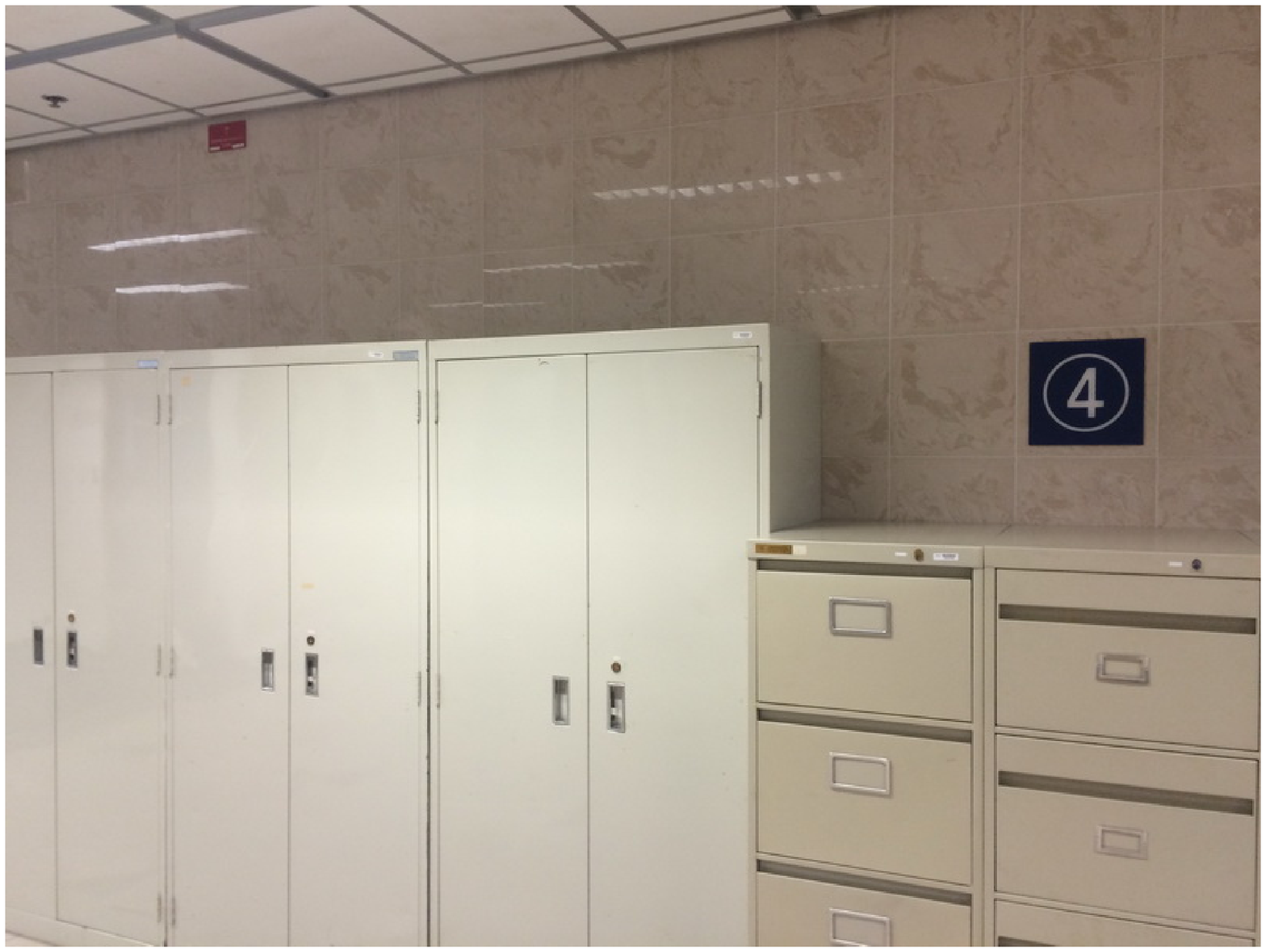}
			\centering{\emph{four}}
		\end{minipage}
		\hfill
		\begin{minipage}[t]{0.24\columnwidth}
			\includegraphics[width=1.0\textwidth]{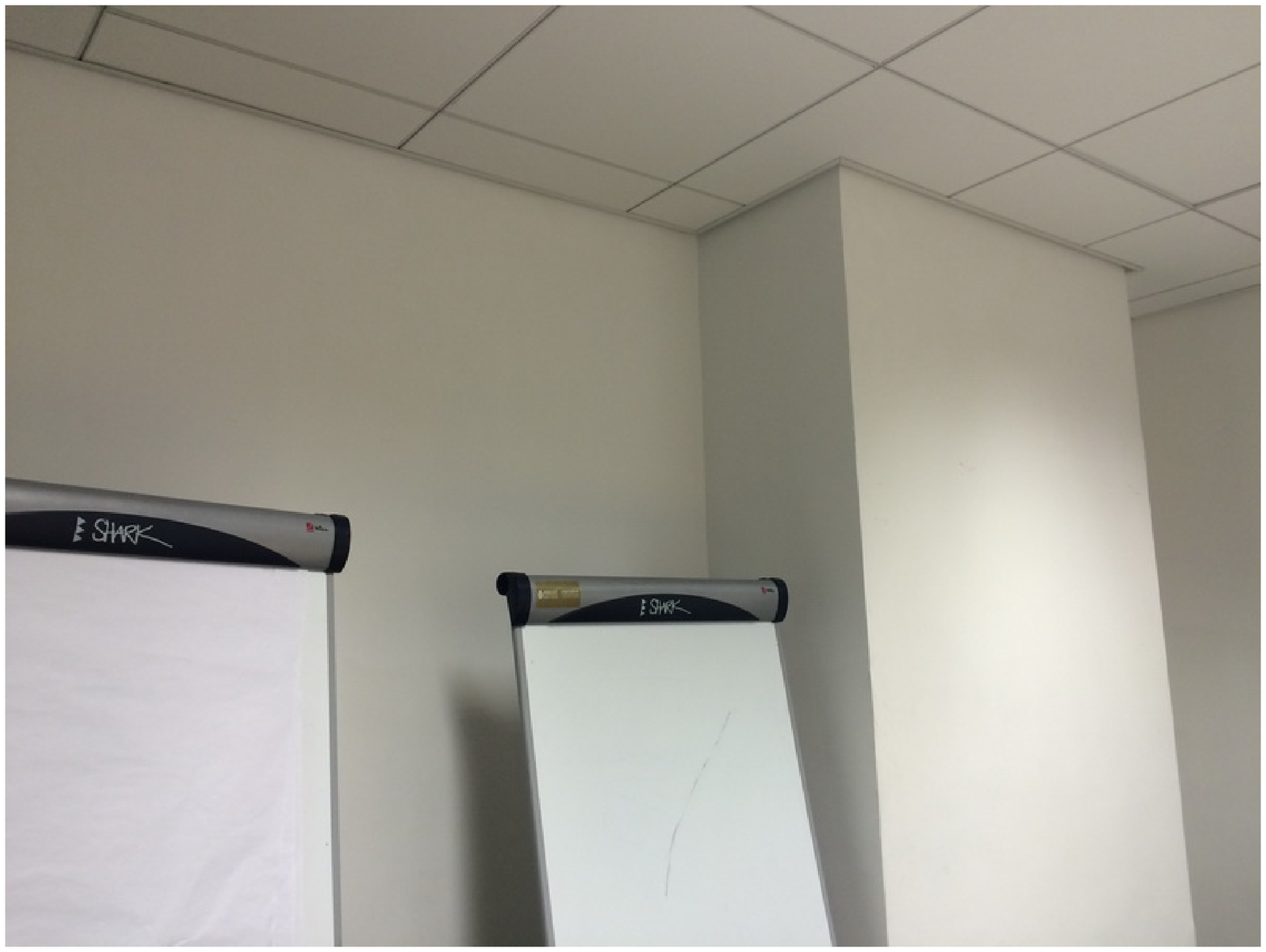}
			\centering{\emph{roof}}
		\end{minipage}
	\end{minipage}
	\vfill
	\begin{minipage}[t]{1.0\columnwidth}
		\begin{minipage}[t]{0.24\columnwidth}
			\includegraphics[width=1.0\textwidth]{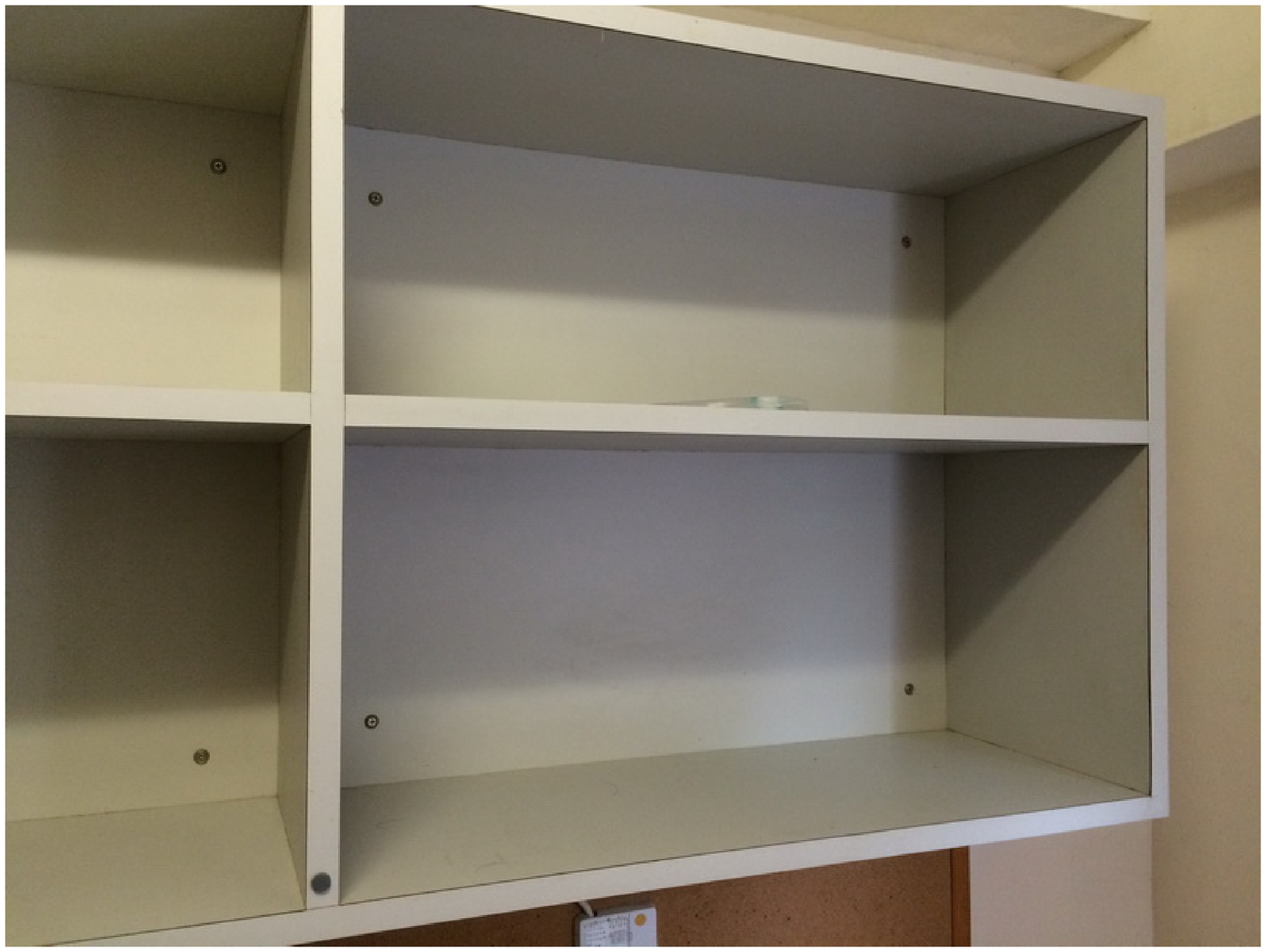}
			\centering{\emph{shelf}}
		\end{minipage}
		\hfill
		\begin{minipage}[t]{0.24\columnwidth}
			\includegraphics[width=1.0\textwidth]{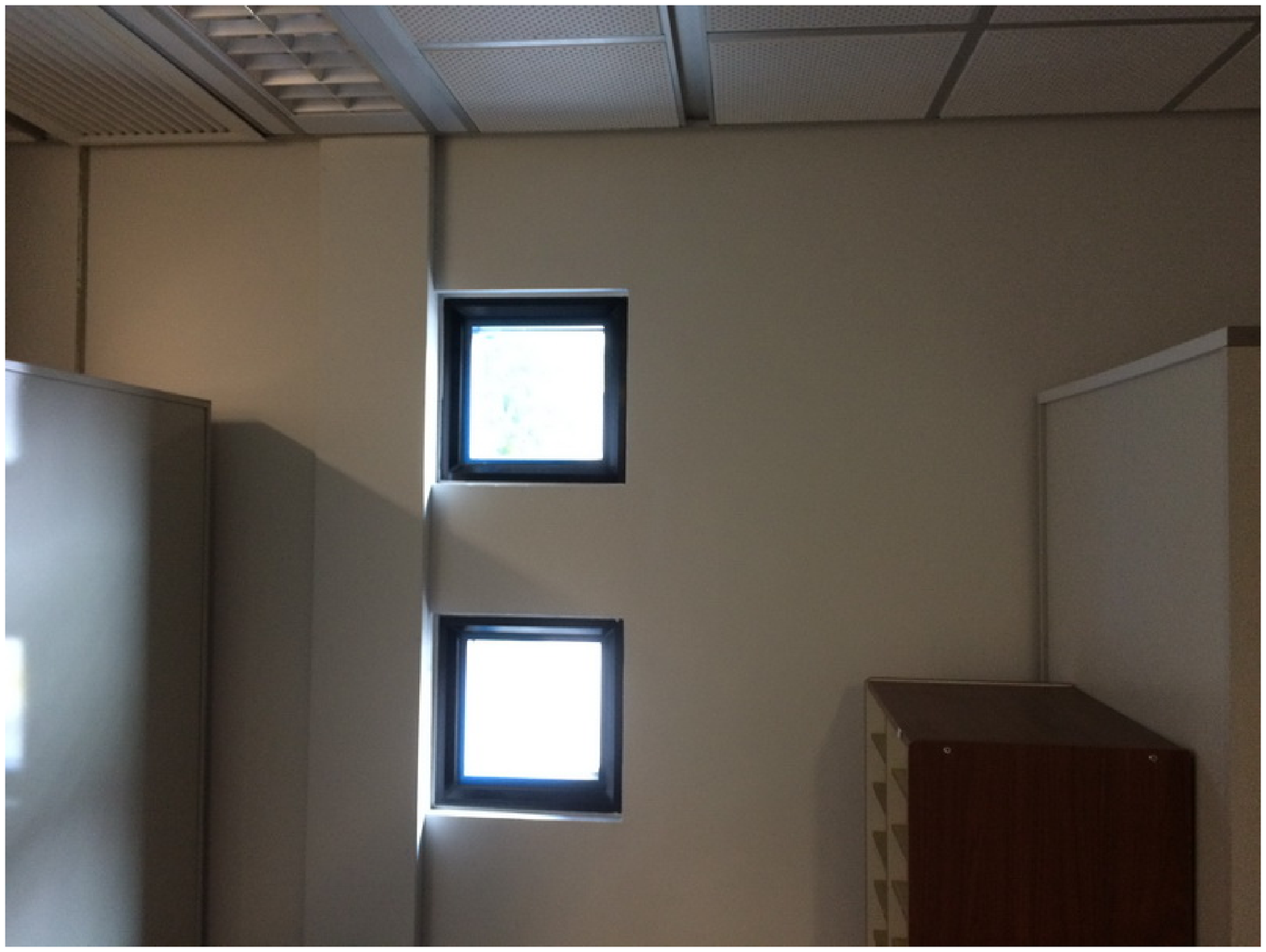}
			\centering{\emph{window}}
		\end{minipage}
		\hfill
		\begin{minipage}[t]{0.24\columnwidth}
			\includegraphics[width=1.0\textwidth]{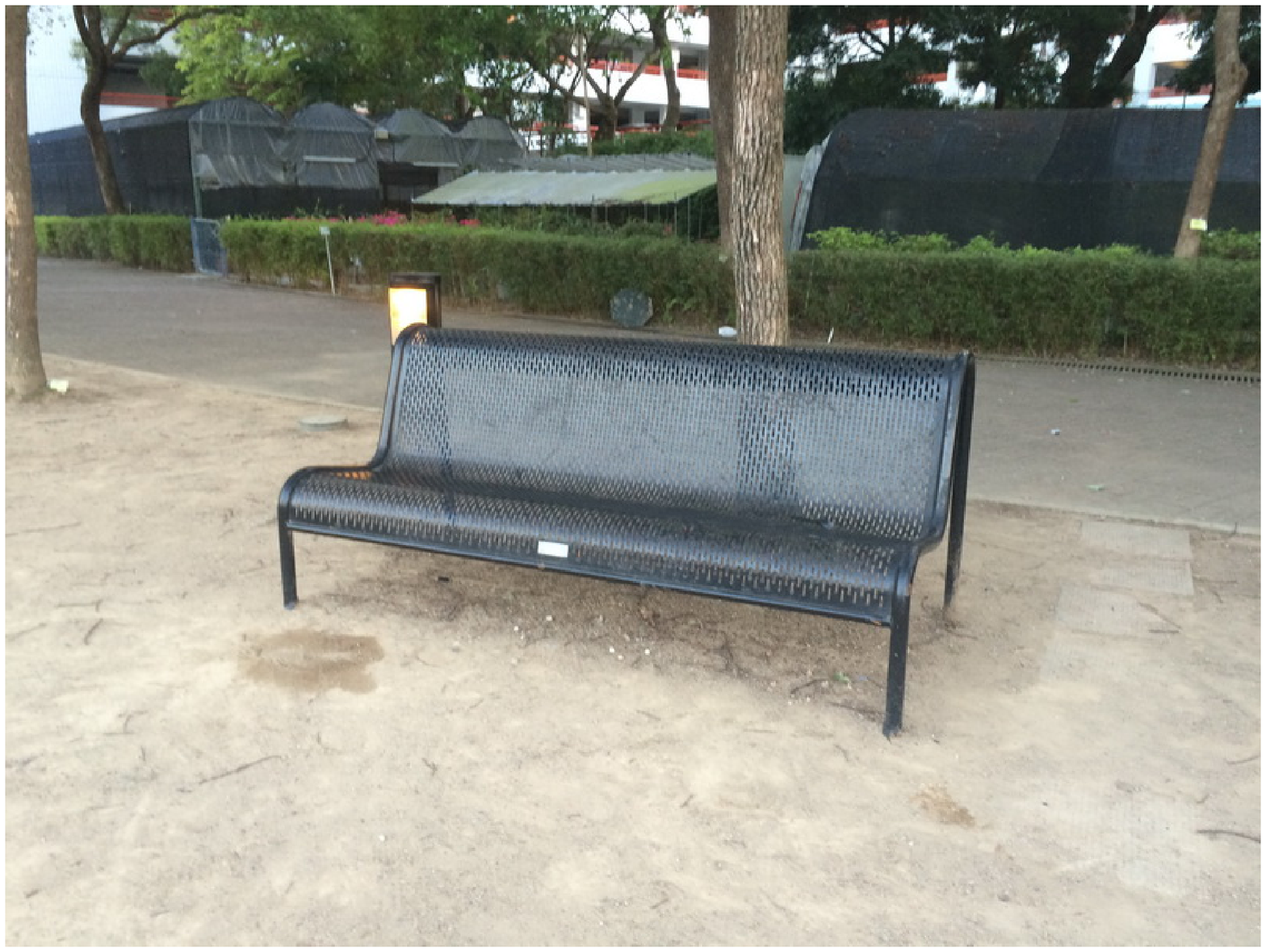}
			\centering{\emph{bench}}
		\end{minipage}
		\hfill
		\begin{minipage}[t]{0.24\columnwidth}
			\includegraphics[width=1.0\textwidth]{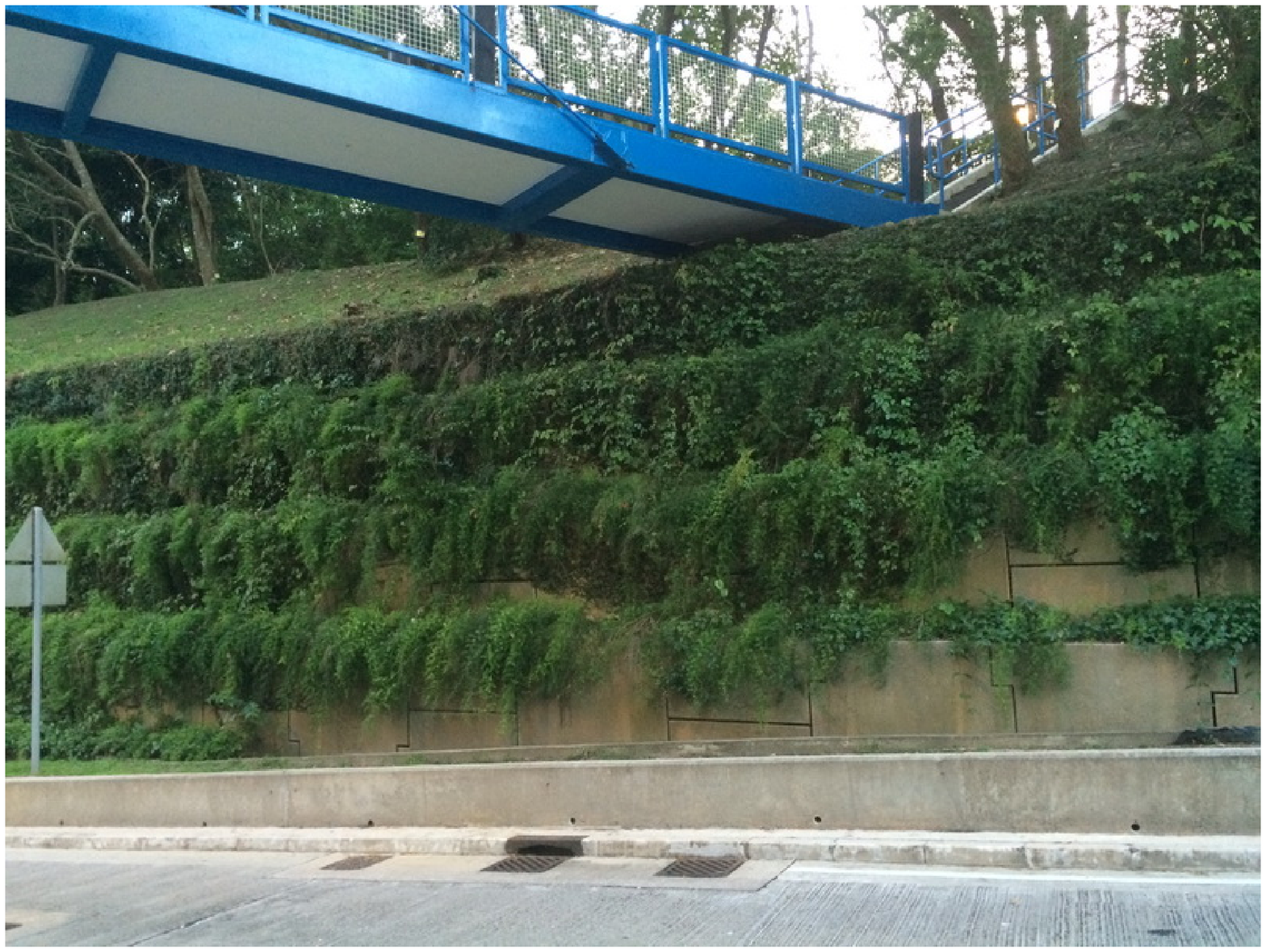}
			\centering{\emph{bridge}}
		\end{minipage}
	\end{minipage}
	\vfill
	\begin{minipage}[t]{1.0\columnwidth}
		\begin{minipage}[t]{0.24\columnwidth}
			\includegraphics[width=1.0\textwidth]{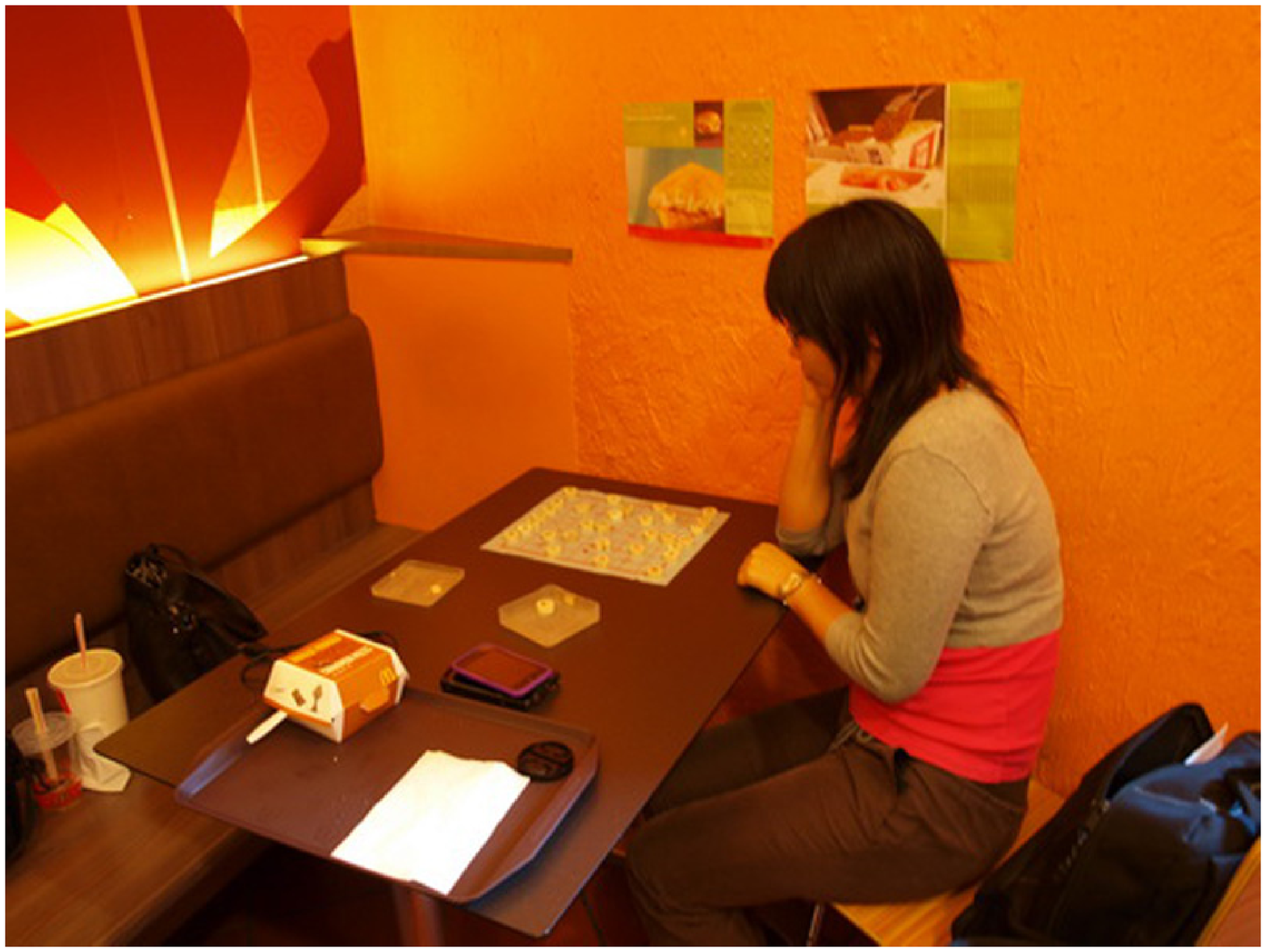}
			\centering{\emph{girl}}
		\end{minipage}
		\hfill
		\begin{minipage}[t]{0.24\columnwidth}
			\includegraphics[width=1.0\textwidth]{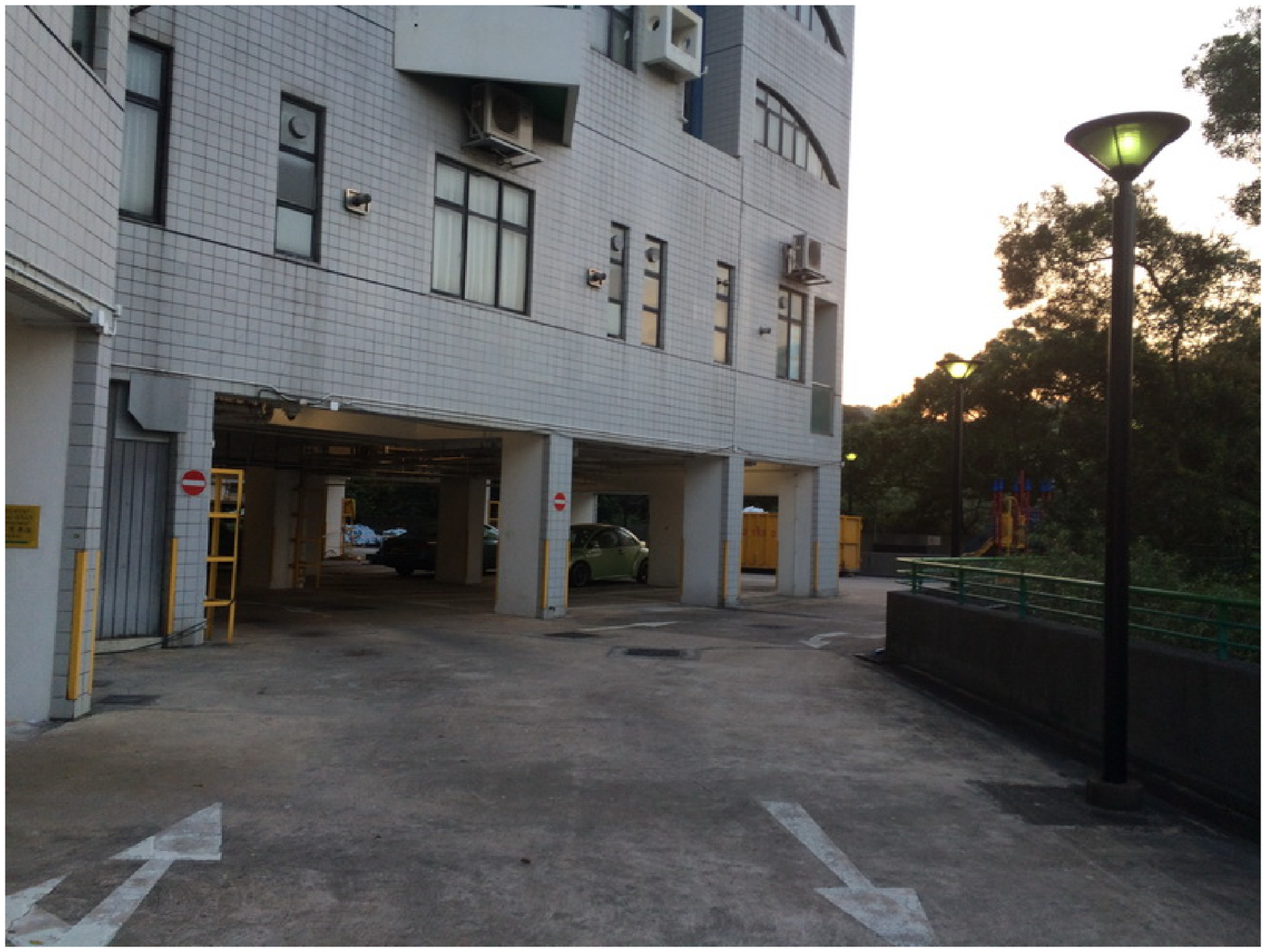}
			\centering{\emph{park}}
		\end{minipage}
		\hfill
		\begin{minipage}[t]{0.24\columnwidth}
			\includegraphics[width=1.0\textwidth]{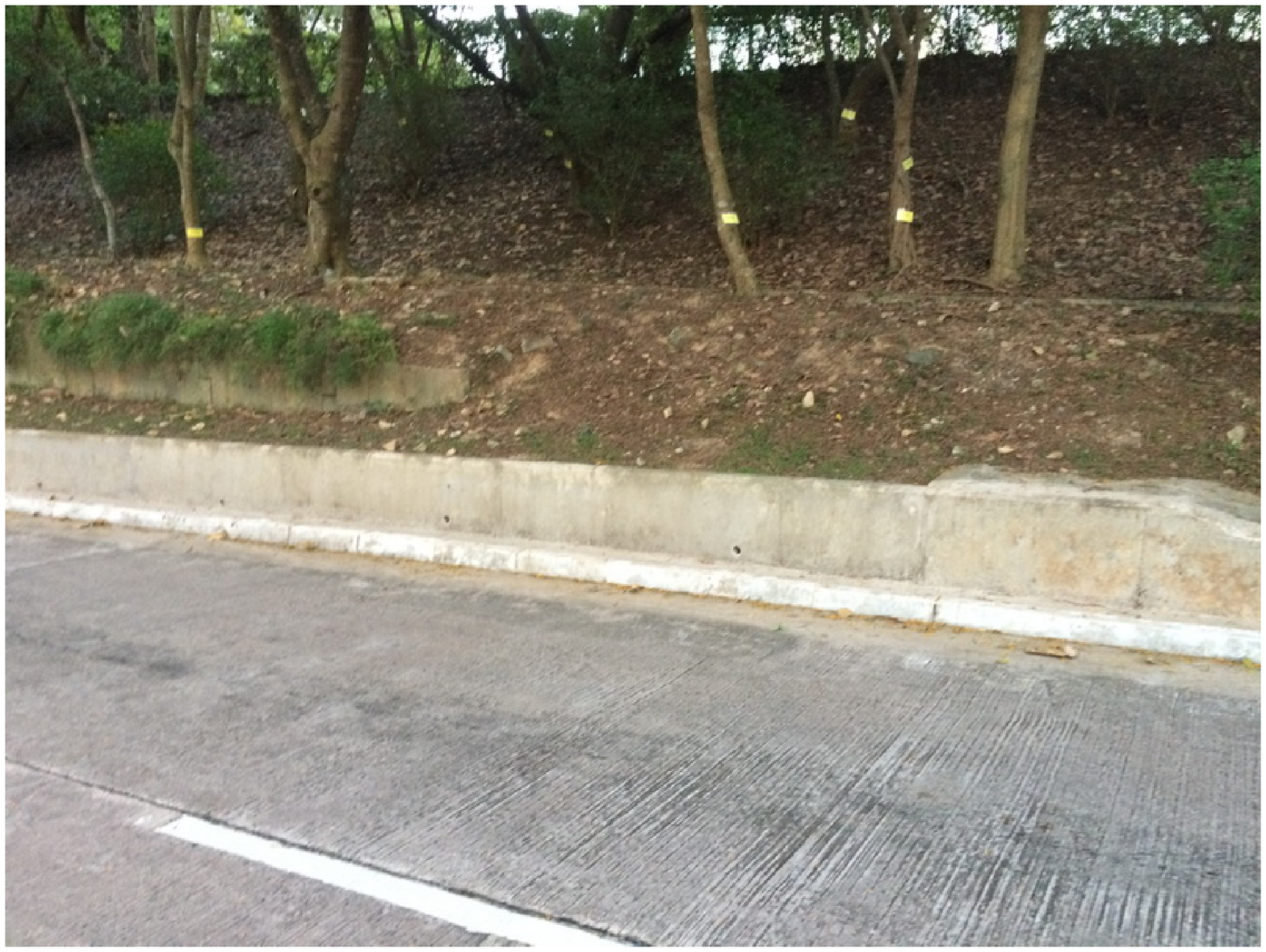}
			\centering{\emph{road}}
		\end{minipage}
		\hfill
		\begin{minipage}[t]{0.24\columnwidth}
			\includegraphics[width=1.0\textwidth]{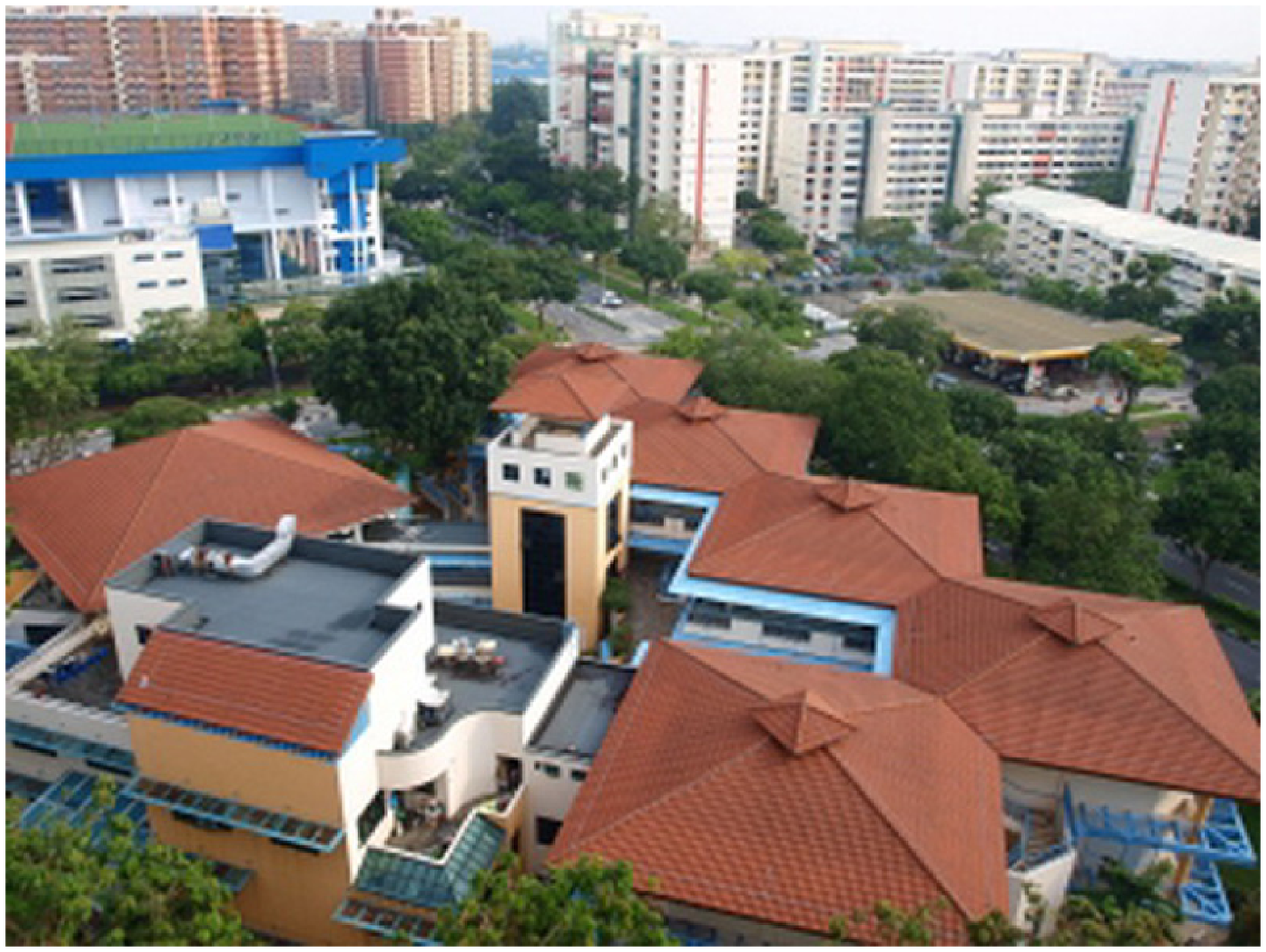}
			\centering{\emph{villa}}
		\end{minipage}
	\end{minipage}
	\caption{The dataset of our comparative experiments on low-texture images.}
	\label{Fig:dataset1}\vspace{-12pt}
\end{figure}

\noindent\textbf{Evaluation on Low-Texture Images.} To demonstrate the performance of our proposed method on stitching images with low textures, as shown in Fig.~\ref{Fig:dataset1}, we evaluate our method on the dataset provided by~\cite{li2015dual}. 

We compare the stitching quality of results produced by APAP, DF-W and our proposed method. For APAP and DF-W, the results are directly obtained from~\cite{li2015dual}. Table.~\ref{Tab:ComResultDFW} presents the comparative results. As we can see, DF-W adopts matching points and lines to constrain the stitching process, which has better results compared with APAP in most cases. Our proposed method combines the photometric constraints with above two geometric constraints and outperforms APAP and DF-W. As for image pair \emph{bench} and \emph{road}, APAP performs better than DF-W because limited line features can be extracted from these two image pairs with wide baseline, while our method still has the best performance because the combination of geometric and photometric constraints.

\begin{table}[t]
	\centering
	\renewcommand{\tabcolsep}{2.5pt}
	\renewcommand\arraystretch{1.05}
	\begin{tabular}{|c|ccc|c|ccc|}
		\hline
		\multicolumn{1}{|c|}{} & \multicolumn{3}{c|}{method} & \multicolumn{1}{c|}{} & \multicolumn{3}{c|}{method}\\ \hline
		Data	&APAP	&DF-W	&Proposed	&Data	&APAP	&DF-W	&Proposed		\\ \hline
		\emph{cabinet} & 4.55 & 2.63 &\textbf{1.33} & \emph{bench} & 4.01 & 7.12 & \textbf{3.64} \\ \hline
		\emph{desk} & 6.17 & 4.89 & \textbf{1.59} & \emph{bridge} & 7.95 & 6.60 & \textbf{4.49} \\ \hline
		\emph{four} & 6.92 & 2.36 & \textbf{0.98} & \emph{girl} & 5.20 & \textbf{4.81} & 5.05 \\ \hline
		\emph{roof} & 7.82 & 2.25 & \textbf{1.52} & \emph{park} & 11.07 & 8.18 & \textbf{5.85} \\ \hline
		\emph{shelf} & 8.76 & 1.54 & \textbf{0.83} & \emph{road} & 2.28 & 4.59 & \textbf{1.67} \\ \hline
		\emph{window} & 5.78 & 4.94 & \textbf{1.95} & \emph{villa} & 6.72 & \textbf{5.20} & 5.41 \\ \hline
	\end{tabular}
	\caption{The RMSE([0,255]) for three compared methods on image pairs. APAP: as-projective-as-possible method~\cite{zaragoza2013projective}; DF-W: dual-feature method~\cite{li2015dual}.}
	\label{Tab:ComResultDFW}\vspace{-10pt}
\end{table}

\begin{figure}[t]
	\centering
	\begin{minipage}[t]{1.0\columnwidth}
		\begin{minipage}[t]{0.24\columnwidth}
			\includegraphics[width=1.0\textwidth]{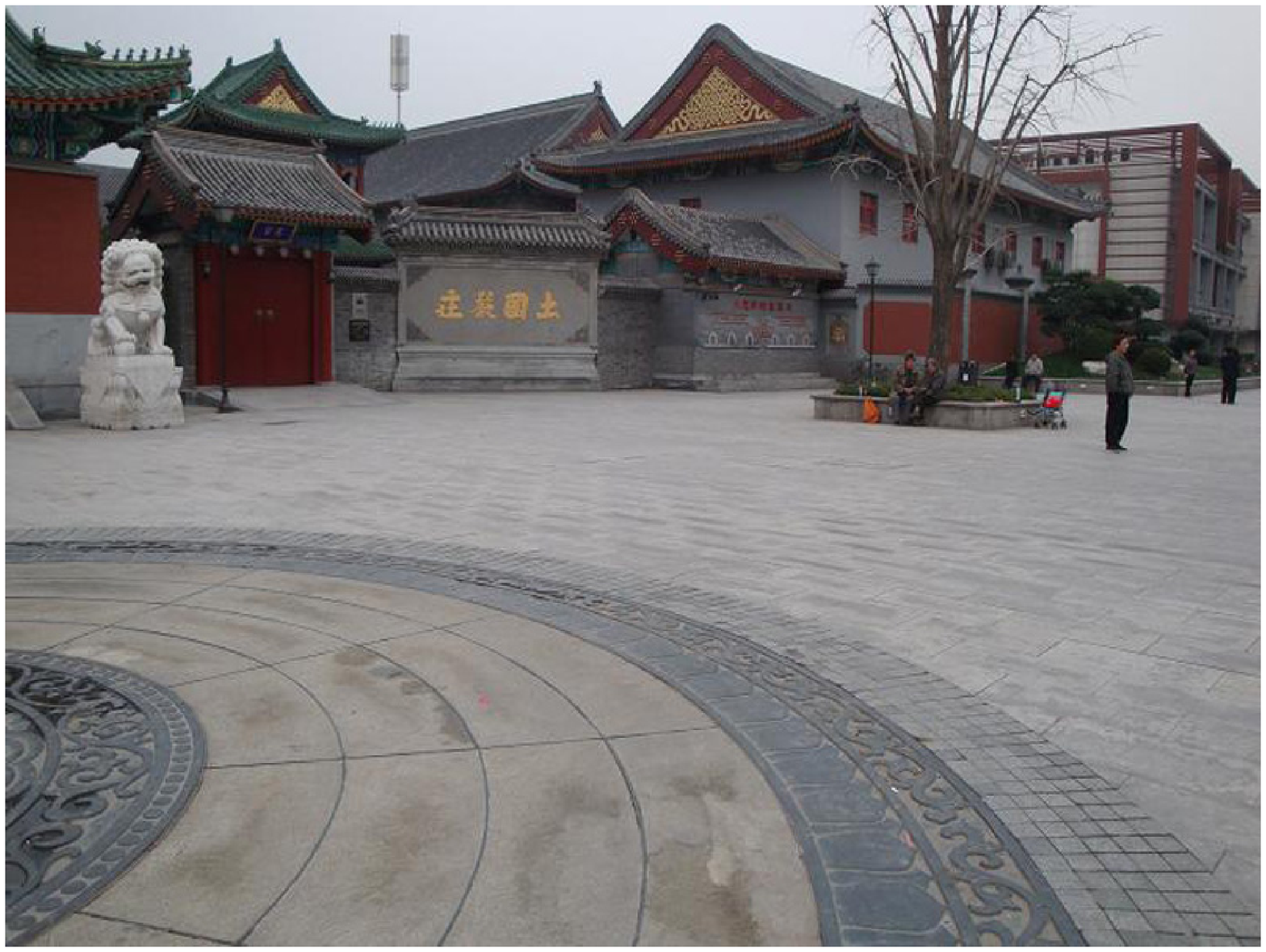}
			\vspace{-1pt}
			\centering{\emph{temple}}
		\end{minipage}
		\hfill
		\begin{minipage}[t]{0.24\columnwidth}
			\includegraphics[width=1.0\textwidth]{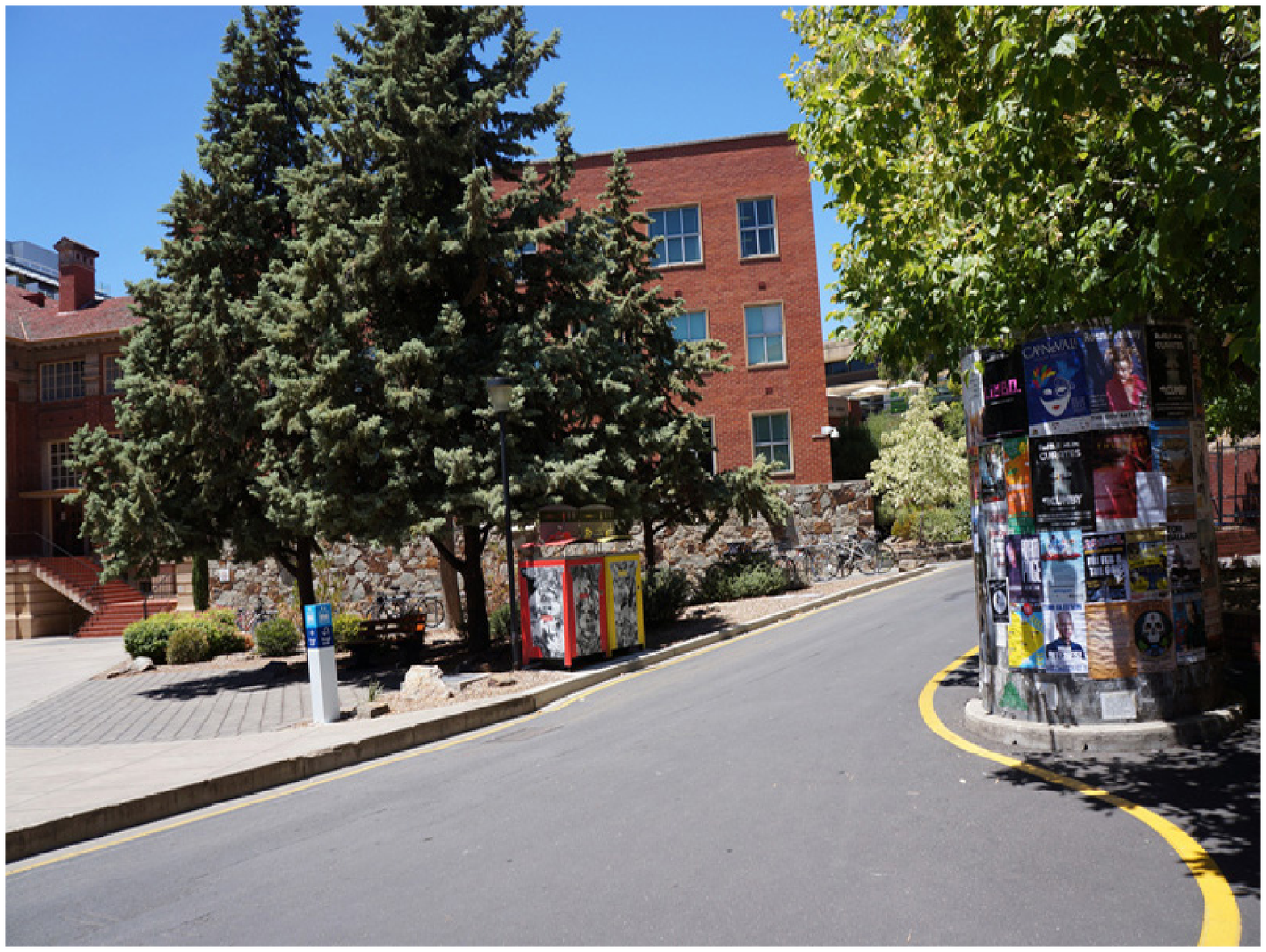}
			\vspace{-1pt}
			\centering{\emph{school}}
		\end{minipage}
		\hfill
		\begin{minipage}[t]{0.24\columnwidth}
			\includegraphics[width=1.0\textwidth]{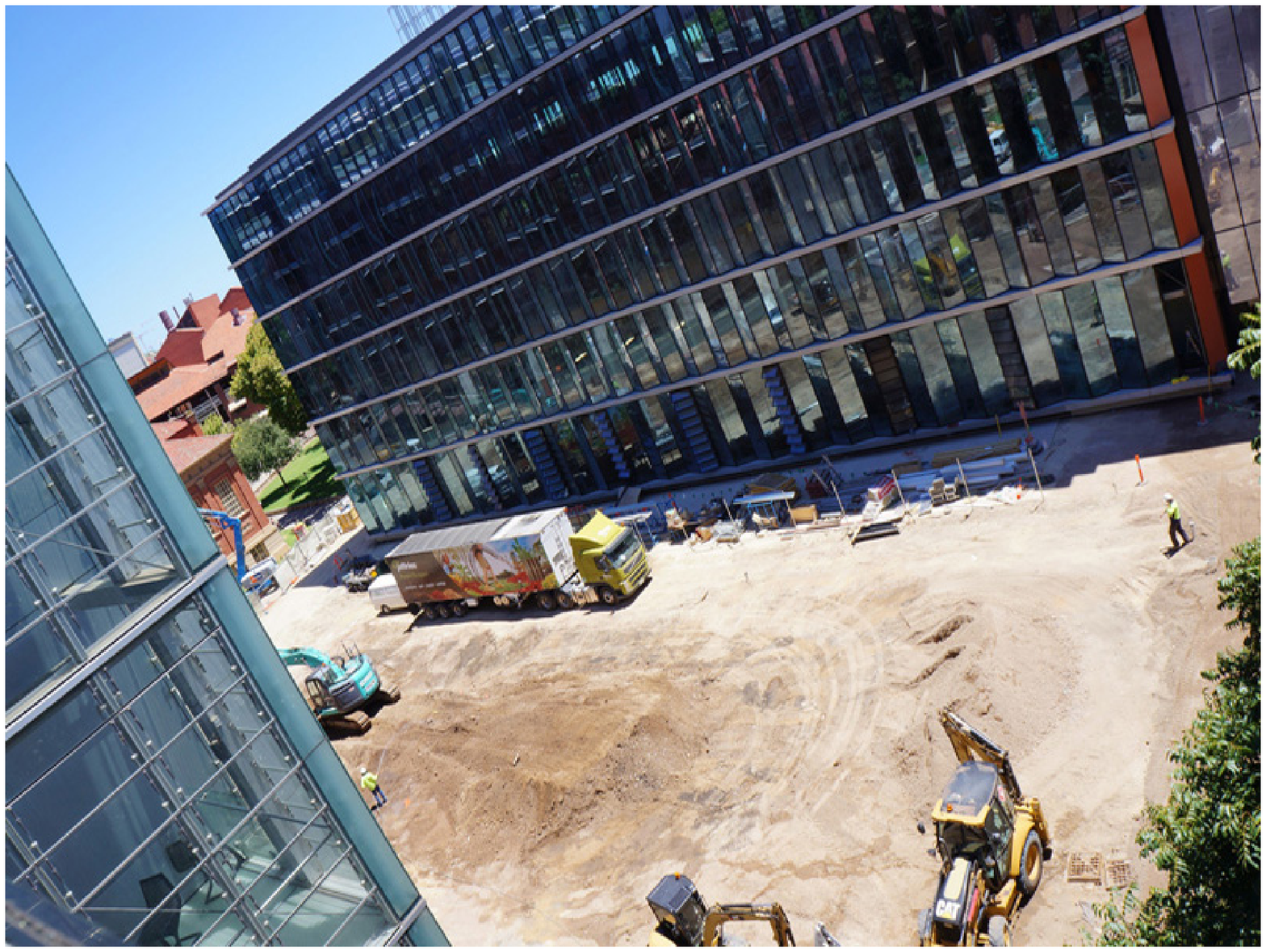}
			\vspace{-1pt}
			\centering{\emph{outdoor}}
		\end{minipage}
		\hfill
		\begin{minipage}[t]{0.24\columnwidth}
			\includegraphics[width=1.0\textwidth]{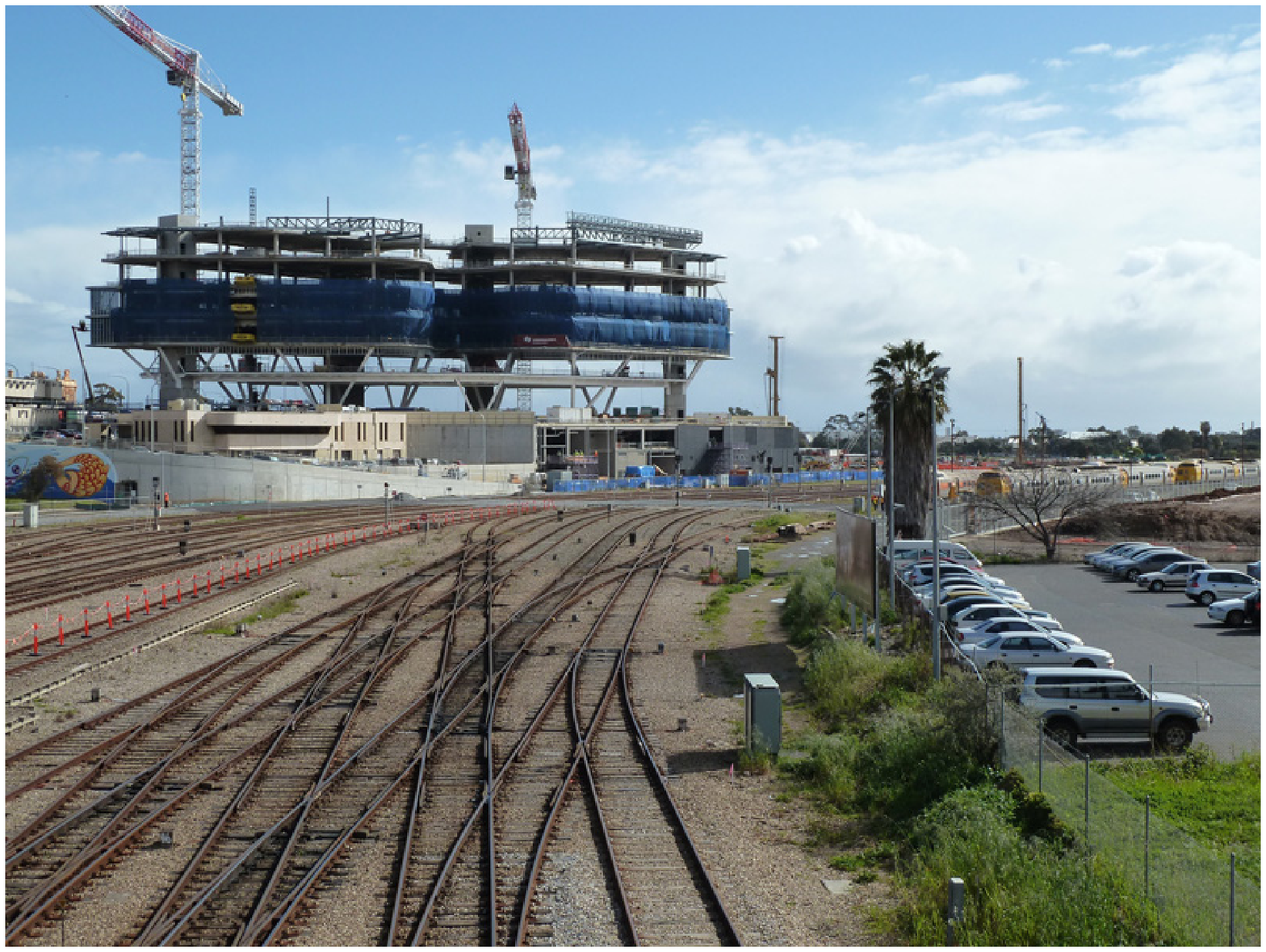}
			\vspace{-1pt}
			\centering{\emph{rail}}
		\end{minipage}
	\end{minipage}
	\vfill
	\vspace{-1pt}
	\begin{minipage}[t]{1.0\columnwidth}
		\begin{minipage}[t]{0.24\columnwidth}
			\includegraphics[width=1.0\textwidth]{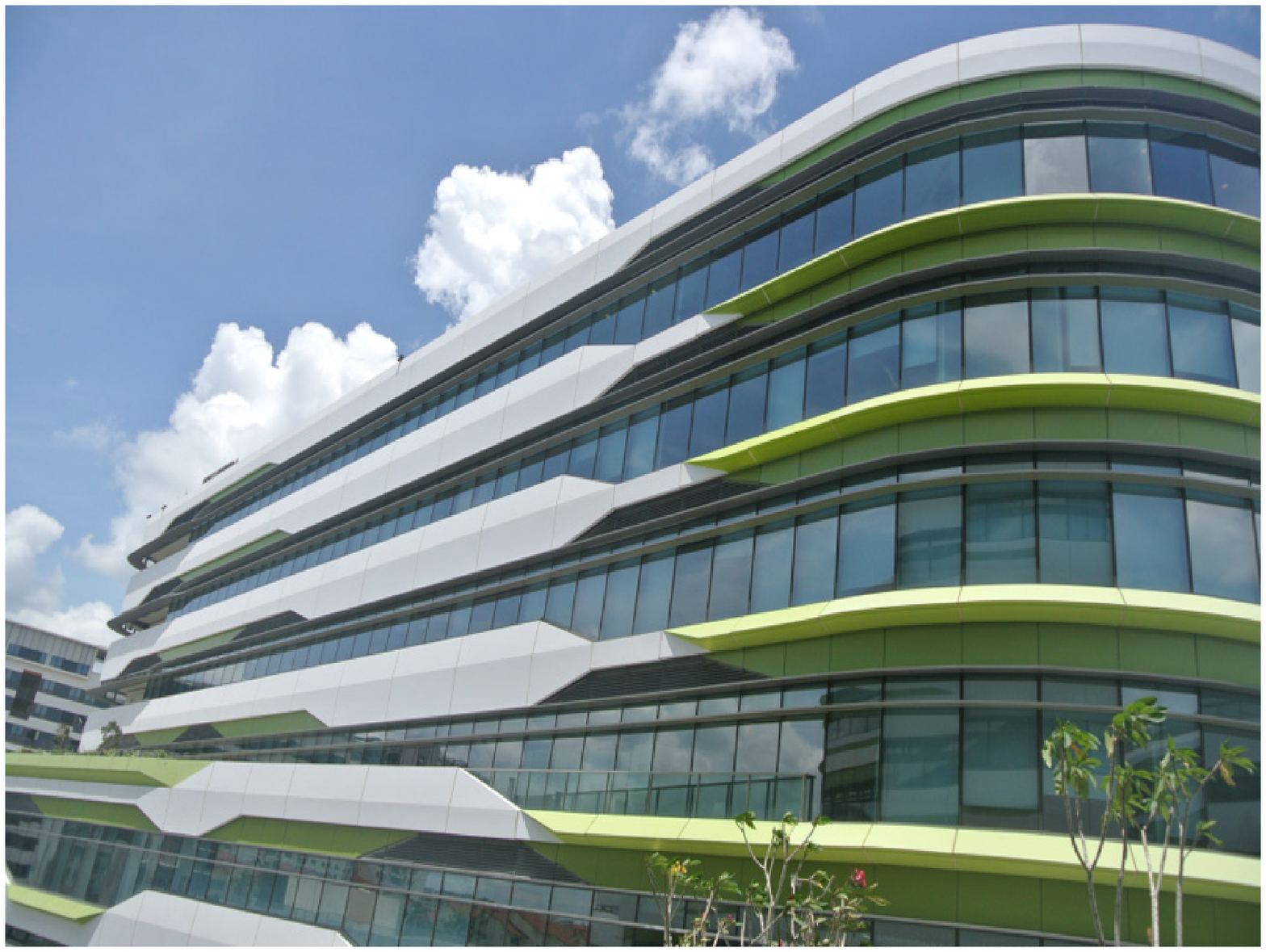}
			\centering{\emph{building}}
		\end{minipage}
		\hfill
		\begin{minipage}[t]{0.24\columnwidth}
			\includegraphics[width=1.0\textwidth]{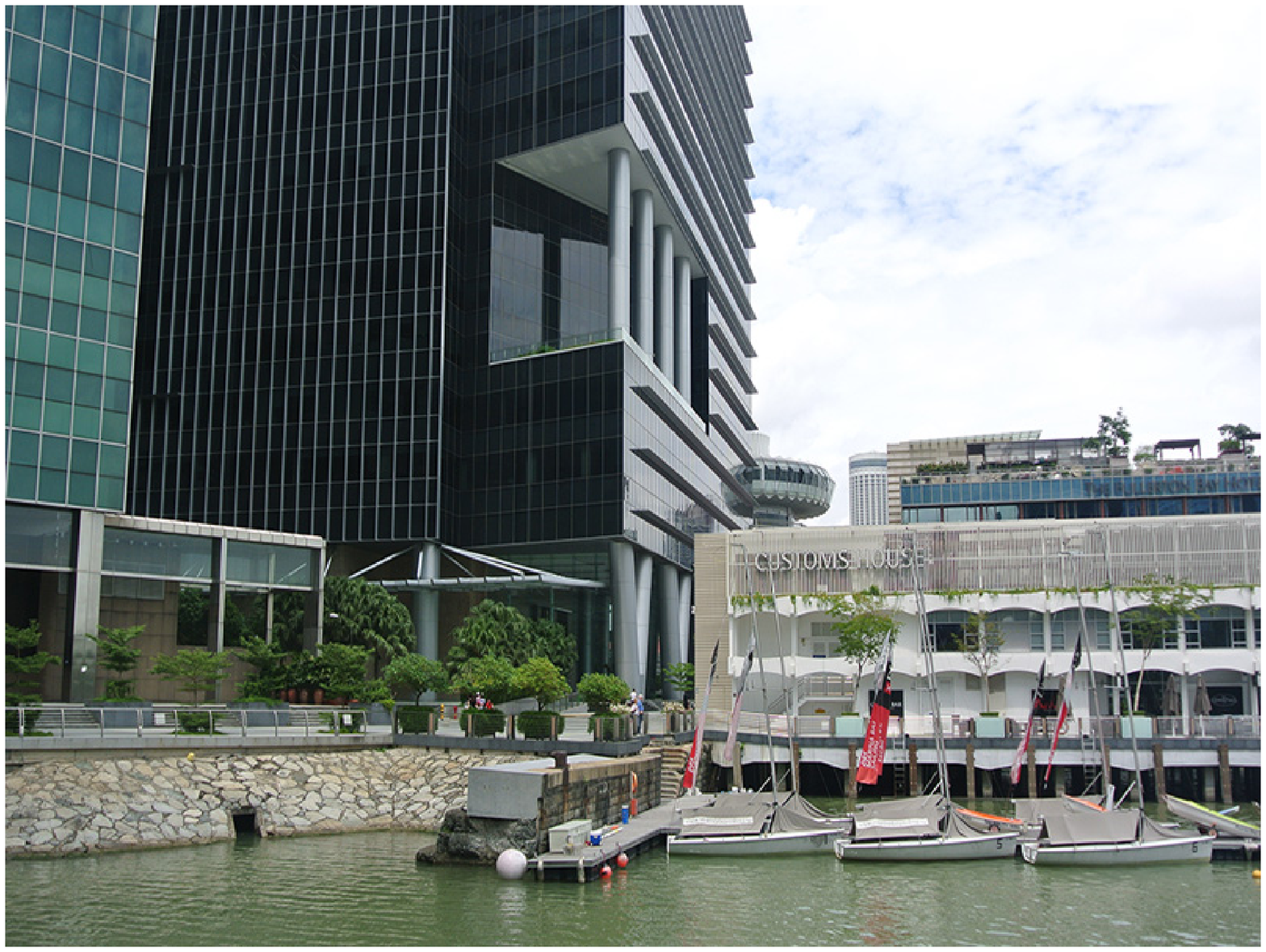}
			\centering{\emph{house}}
		\end{minipage}
		\hfill
		\begin{minipage}[t]{0.24\columnwidth}
			\includegraphics[width=1.0\textwidth]{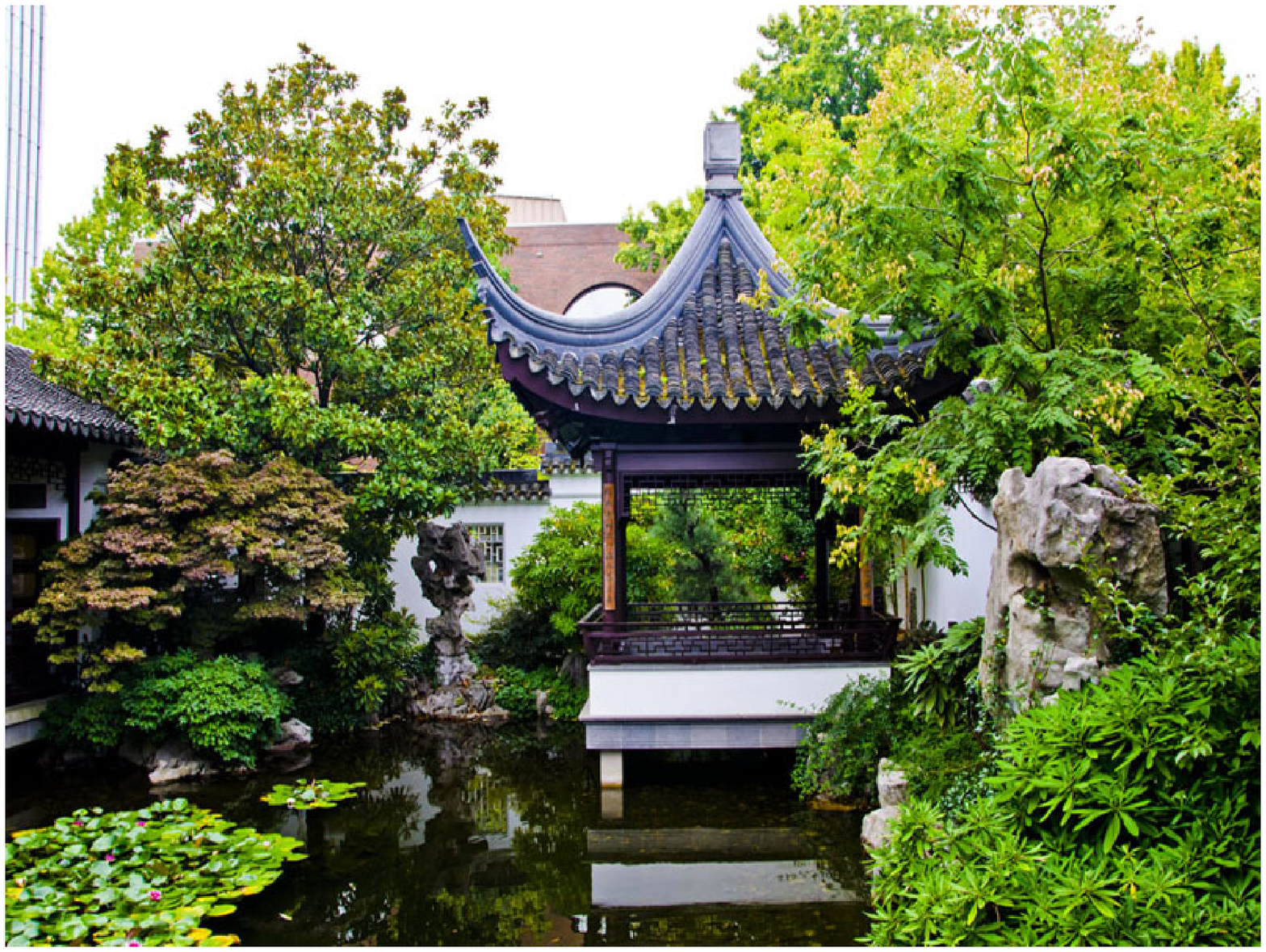}
			\centering{\emph{courtyard}}
		\end{minipage}
		\hfill
		\begin{minipage}[t]{0.24\columnwidth}
			\includegraphics[width=1.0\textwidth]{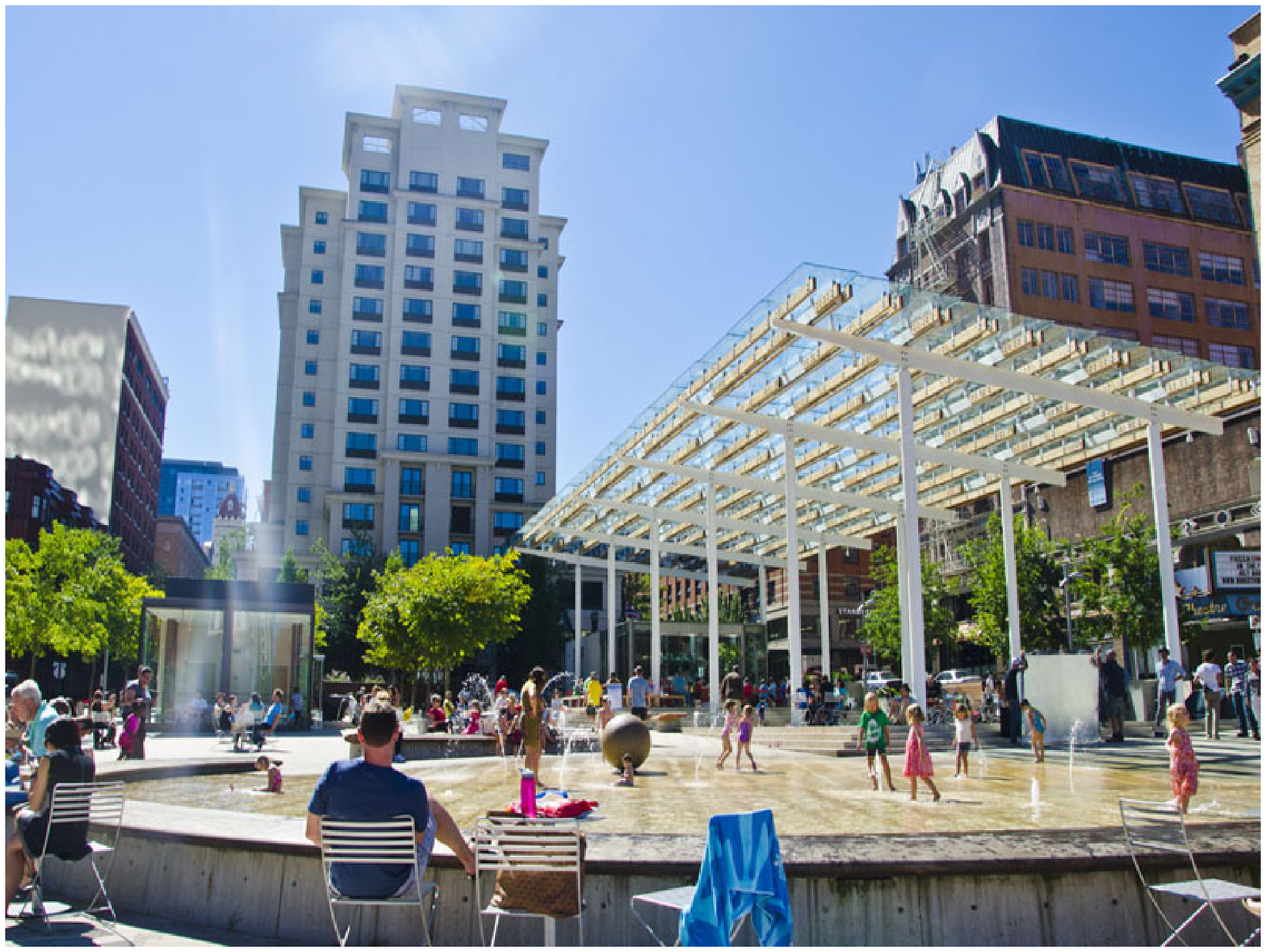}
			\centering{\emph{square}}
		\end{minipage}
	\end{minipage}
	\vspace{-12pt}
	\caption{The dataset of comparative experiments on large-parallax images.}
	\label{Fig:dataset2}\vspace{-10pt}
\end{figure}

\vspace{-7pt}
\begin{table}[t]
	\centering
	\vspace{3pt}
	\renewcommand{\tabcolsep}{10pt}
	\renewcommand\arraystretch{1.05}
	\begin{tabular}{|c|c|c|c|c|}
		\hline
		Data &APAP &MPA &Initial &Proposed \\ \hline
		\emph{temple} &6.39 &4.65 &5.56 &\textbf{2.57} \\ \hline
		\emph{school} &12.20 &\textbf{9.73} &22.83 &10.85 \\ \hline
		\emph{outdoor} &11.90 &10.40 &12.52 &\textbf{6.75} \\ \hline
		\emph{rail} &14.80 &11.80 &21.34 &\textbf{9.81} \\ \hline
		\emph{building} &6.68 &4.94 &6.48 &\textbf{3.74} \\ \hline
		\emph{house} &19.80 &18.00 &19.72 &\textbf{14.57} \\ \hline
		\emph{courtyard} &38.30 &32.50 &41.23 &\textbf{29.17} \\ \hline
		\emph{square} &19.90 &16.80 &16.23 &\textbf{12.55} \\ \hline
	\end{tabular}
	\vspace{-5pt}
	\caption{The RMSE result compared with APAP and MPA on image pairs with large parallax. MPA: Mesh-based photometric alignment method~\cite{lin2017direct}. Initial: Our global alignment method.}
	\label{Tab:ComResultMPA}\vspace{-18pt}
\end{table}

\begin{figure*}[t]
	\centering
	\begin{minipage}[t]{1.0\textwidth}
		\begin{minipage}[t]{0.49\textwidth}
			\includegraphics[width=1.0\textwidth]{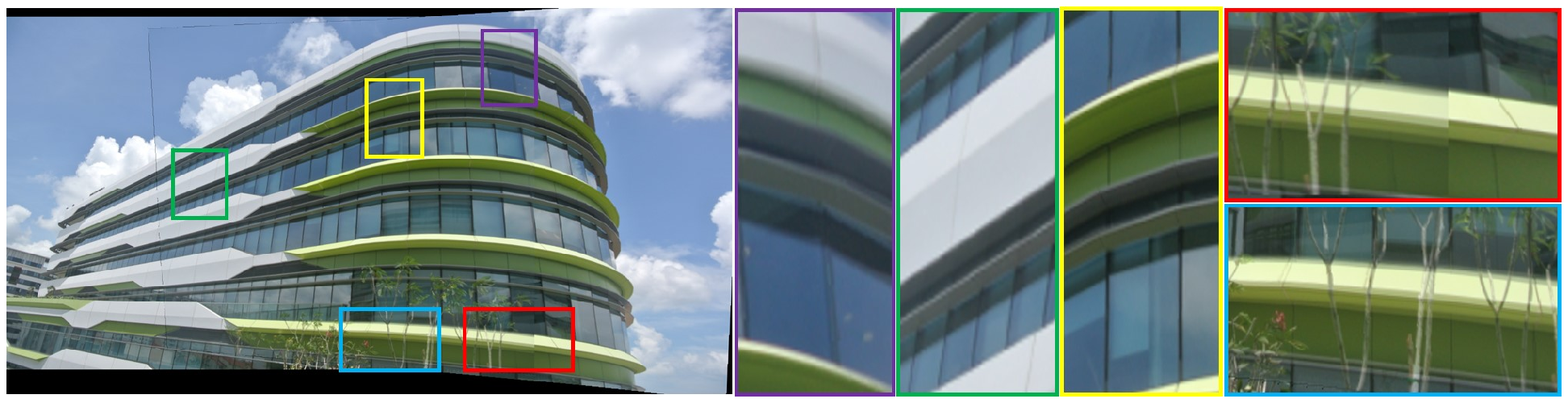}
		\end{minipage}
		\hfill
		\begin{minipage}[t]{0.49\textwidth}
			\includegraphics[width=1.0\textwidth]{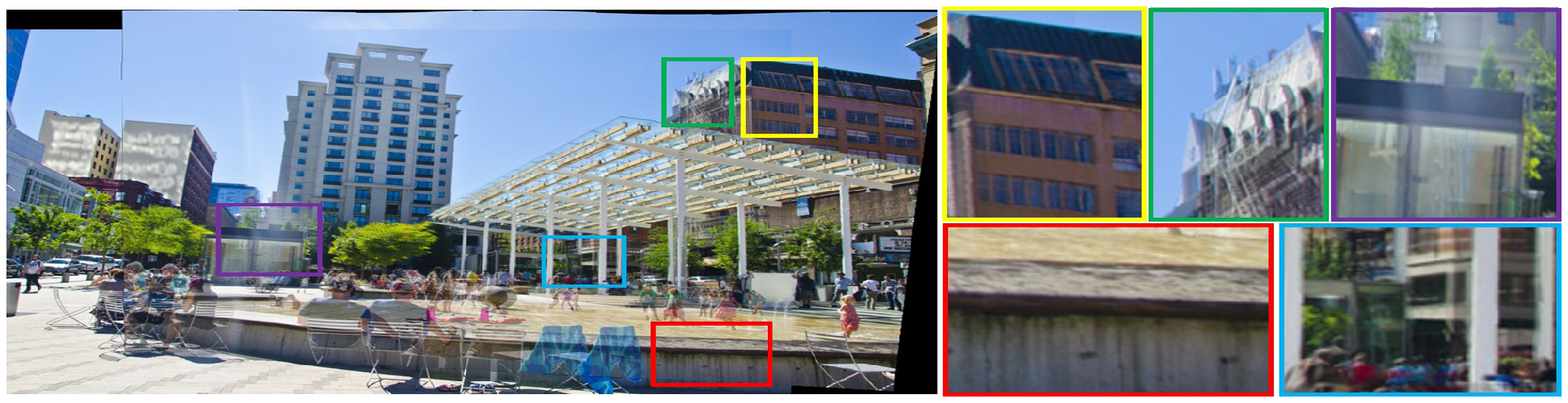}
		\end{minipage}
	\end{minipage}
	\vfill
	\begin{minipage}[t]{1.0\textwidth}
		\begin{minipage}[t]{0.49\textwidth}
			\includegraphics[width=1.0\textwidth]{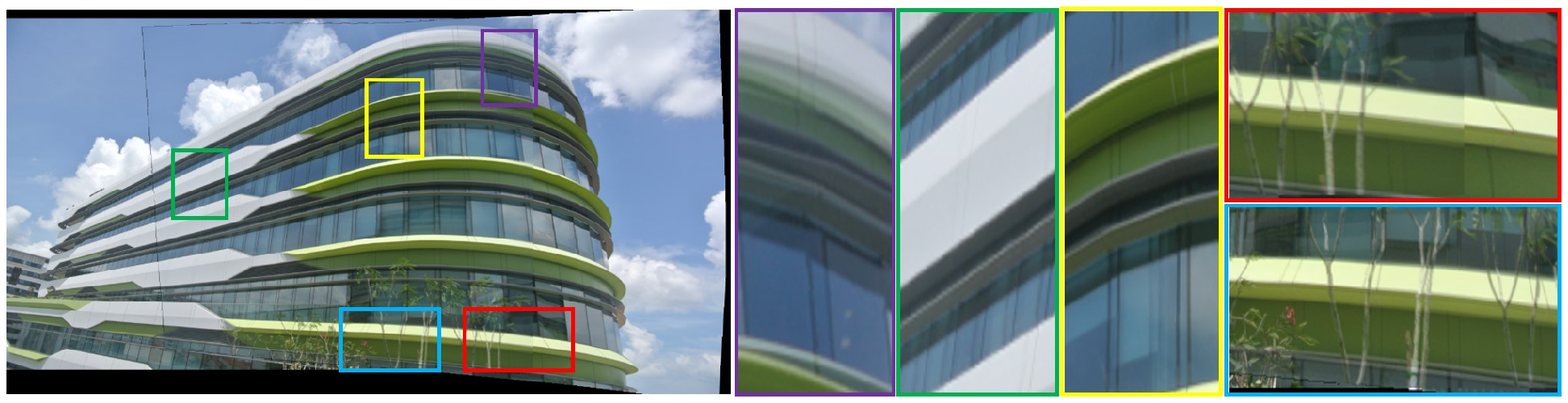}
		\end{minipage}
		\hfill
		\begin{minipage}[t]{0.49\textwidth}
			\includegraphics[width=1.0\textwidth]{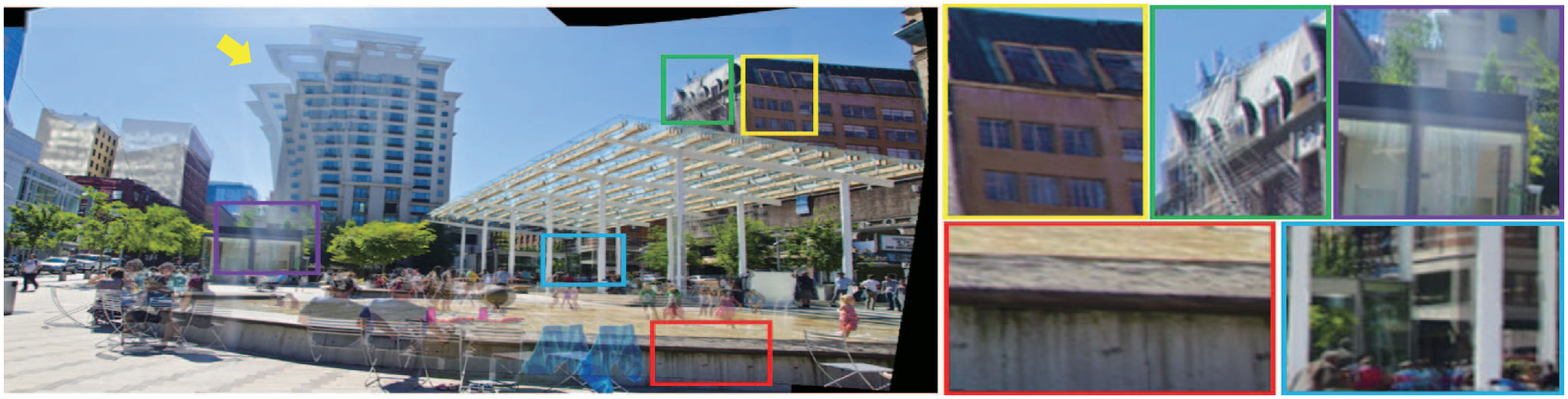}
		\end{minipage}
	\end{minipage}
	\vfill
	\begin{minipage}[t]{1.0\textwidth}
		\begin{minipage}[t]{0.49\textwidth}
			\includegraphics[width=1.0\textwidth]{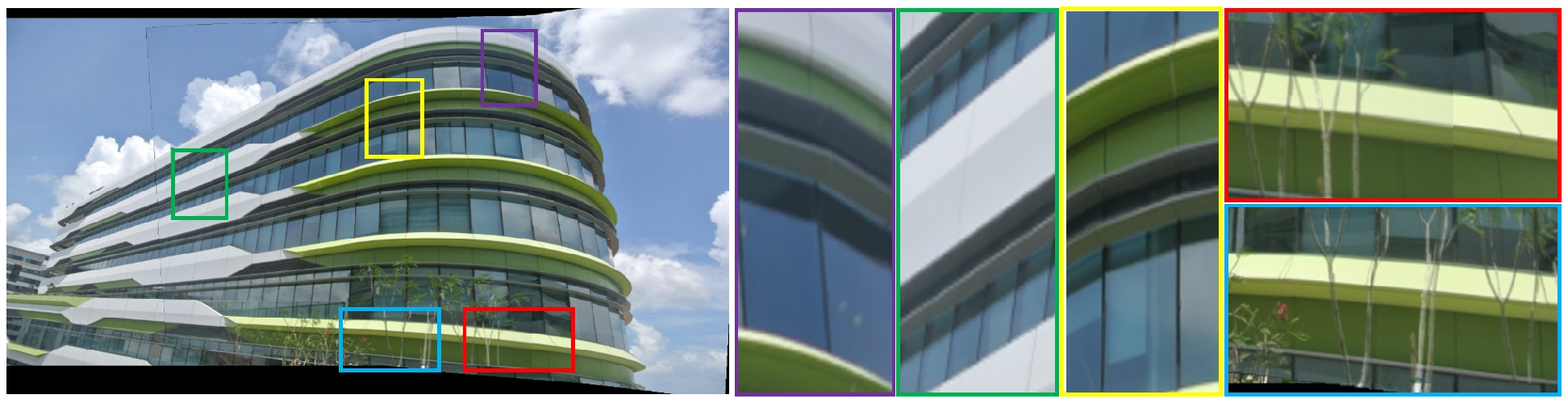}
		\end{minipage}
		\hfill
		\begin{minipage}[t]{0.49\textwidth}
			\includegraphics[width=1.0\textwidth]{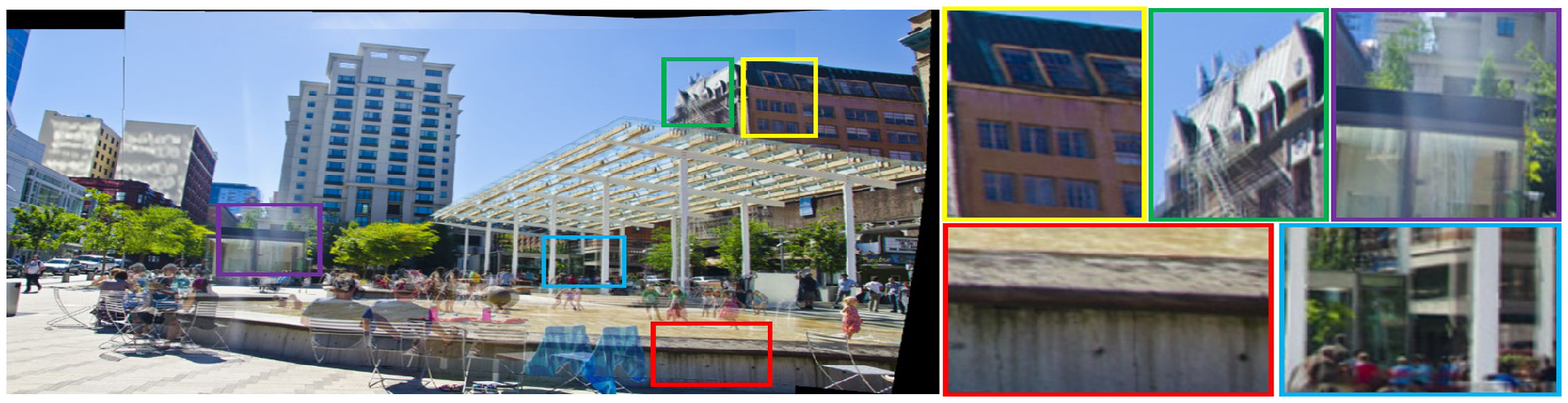}
		\end{minipage}
	\end{minipage}
	\vspace{-18pt}
	\caption{Qualitative comparisons between between methods with three different constraint combinations. \textbf{First Row:} Results with geometric constraints; \textbf{Second Row:} Results with geometric constraints followed by photometric constraints; \textbf{Third Row:} Results with geometric constraints combined with photometric constraints.}
	\label{Fig:Qualitative}\vspace{-15pt}
\end{figure*}

\vspace{5pt}
\noindent\textbf{Evaluation on Large-Parallax Images.} We evaluate the performance of our proposed method on stitching large-parallax images on the dataset presented in Fig.~\ref{Fig:dataset2}. It consists of $8$ pairs of images with large parallaxes, which are collected from publicly available datasets~\cite{zaragoza2013projective,zhang2014parallax,lin2016seagull}.

We compare the alignment accuracy of results produced by APAP, MPA~\cite{lin2017direct} and our proposed method. Table.~\ref{Tab:ComResultMPA} shows the comparative results. MPA uses dense photometric constraints to restrain a mesh-based stitching model and gets better alignment accuracy than APAP. However, similar to the problem of optical flow estimation, optimization with photometric constraint is easily trapped into local optimums and may not work well on images with large parallax. Our proposed method combines photometric constraints with geometric constraints to handle large parallax and usually has lower stitching errors than MPA.

\vspace{-12pt}
\subsection{Qualitative Comparison}
\label{Subsec:Qualitative}
\vspace{-7pt}

We design a comparative experiment to demonstrate the superiority of our proposed method by comparing stitching results produced by methods with three different constraint combinations: (1) geometric constraints; (2) geometric constraints followed by photometric constraints; (3) geometric constraints combined with photometric constraints. It is to be noted that the second combination is different from the third one (our proposed). The second combination imposes photometric constraints after geometric constraints and refers to the photometric item as a post-processing tool, which is similar to~\cite{lin2017direct}. 

Fig.~\ref{Fig:Qualitative} presentsthe qualitative results. Mesh-based warping with point and line constraints fail to get satisfactory results on two selected image pairs where the extraction and matching of line segments are difficult for complex image structures. The following photometric-based post-process can improve the alignment quality locally but the improvements are very limited. As the photometric-based optimization is sensitive to initial values and is easily trapped into local minimum, in our experiment, this post-process sometimes may deteriorate the stitching result. In contrast, the results obtained by our proposed method always have good alignment quality in the overlapping region, because photometric constraints make a complement to geometric constraints in the joint optimization process.

\vspace{-14pt}
\section{Conclusion}
\label{Sec:Conclusion}
\vspace{-10pt}

In this paper, we propose a mesh-based warping method for image stitching, which combines the geometric constraints and the photometric constraints. The photometric constraints can always extracted from image content evenly and densely, and the geometric constraints are more robust to handle large parallax. Jointly utilize these two kinds of constraints in an unified content-preserving warping framework makes them complementary. A lot of quantitative and qualitative experiments demonstrate the superiority of our proposed image stitching method.

\vspace{-10pt}
\section*{\small Acknowledgement}
\vspace{-10pt}
\thanks \small{This work was partially supported by the National Natural Science Foundation of China (Project No. 41571436), the National Key Research and Development Program of China (Project No. 2017YFB1302400), the Hubei Province Science and Technology Support Program, China (Project No. 2015BAA027), the National Natural Science Foundation of China under Grant 91438203, and LIESMARS  Special Research Funding.}


\newpage
\bibliographystyle{IEEEbib}
\bibliography{refs}

\begin{thebibliography}{10}

\bibitem{brown2007automatic}
Matthew Brown and David~G Lowe,
\newblock ``Automatic panoramic image stitching using invariant features,''
\newblock {\em International Journal of Computer Vision}, vol. 74, no. 1, pp.
  59--73, 2007.

\bibitem{gao2011constructing}
Junhong Gao, Seon~Joo Kim, and Michael~S Brown,
\newblock ``Constructing image panoramas using dual-homography warping,''
\newblock in {\em IEEE Conference on Computer Vision and Pattern Recognition
  (CVPR)}, 2011.

\bibitem{lin2011smoothly}
Wen-Yan Lin, Siying Liu, Yasuyuki Matsushita, Tian-Tsong Ng, and Loong-Fah
  Cheong,
\newblock ``Smoothly varying affine stitching,''
\newblock in {\em IEEE Conference on Computer Vision and Pattern Recognition
  (CVPR)}, 2011.

\bibitem{zaragoza2013projective}
Julio Zaragoza, Tat-Jun Chin, Michael~S Brown, and David Suter,
\newblock ``As-projective-as-possible image stitching with moving dlt,''
\newblock in {\em IEEE Conference on Computer Vision and Pattern Recognition
  (CVPR)}, 2013.

\bibitem{chang2014shape}
Che-Han Chang, Yoichi Sato, and Yung-Yu Chuang,
\newblock ``Shape-preserving half-projective warps for image stitching,''
\newblock in {\em IEEE Conference on Computer Vision and Pattern Recognition
  (CVPR)}, 2014.

\bibitem{chai2016shape}
Qingpeng Chai and Shiguang Liu,
\newblock ``Shape-optimizing hybrid warping for image stitching,''
\newblock in {\em IEEE International Conference on Multimedia and Expo (ICME)},
  2016.

\bibitem{zhang2014parallax}
Fan Zhang and Feng Liu,
\newblock ``Parallax-tolerant image stitching,''
\newblock in {\em IEEE Conference on Computer Vision and Pattern Recognition
  (CVPR)}, 2014.

\bibitem{joo2015line}
Kyungdon Joo, Namil Kim, Tae-Hyun Oh, and In~So Kweon,
\newblock ``Line meets as-projective-as-possible image stitching with moving
  dlt,''
\newblock in {\em IEEE International Conference on Image Processing (ICIP)},
  2015.

\bibitem{chen2016natural}
Yu-Sheng Chen and Yung-Yu Chuang,
\newblock ``Natural image stitching with the global similarity prior,''
\newblock in {\em European Conference on Computer Vision (ECCV)}, 2016.

\bibitem{hu2015multi}
Jie Hu, Dong-Qing Zhang, Heather Yu, and Chang~Wen Chen,
\newblock ``Multi-objective content preserving warping for image stitching,''
\newblock in {\em IEEE International Conference on Multimedia and Expo (ICME)},
  2015.

\bibitem{li2015dual}
Shiwei Li, Lu~Yuan, Jian Sun, and Long Quan,
\newblock ``Dual-feature warping-based motion model estimation,''
\newblock in {\em IEEE International Conference on Computer Vision (ICCV)},
  2015.

\bibitem{lin2016seagull}
Kaimo Lin, Nianjuan Jiang, Loong-Fah Cheong, Minh Do, and Jiangbo Lu,
\newblock ``Seagull: Seam-guided local alignment for parallax-tolerant image
  stitching,''
\newblock in {\em European Conference on Computer Vision (ECCV)}, 2016.

\bibitem{xiang2016locally}
Tianzhu Xiang, Gui-Song Xia, Liangpei Zhang, and Ningning Huang,
\newblock ``Locally warping-based image stitching by imposing line
  constraints,''
\newblock in {\em International Conference on Pattern Recognition (ICPR)},
  2016.

\bibitem{zhang2016multi}
Guofeng Zhang, Yi~He, Weifeng Chen, Jiaya Jia, and Hujun Bao,
\newblock ``Multi-viewpoint panorama construction with wide-baseline images,''
\newblock {\em IEEE Transactions on Image Processing}, vol. 25, no. 7, pp.
  3099--3111, 2016.

\bibitem{lin2017direct}
Kaimo Lin, Nianjuan Jiang, Shuaicheng Liu, Loong-Fah Cheong, and Minh
  Do2~Jiangbo Lu,
\newblock ``Direct photometric alignment by mesh deformation,''
\newblock in {\em IEEE Conference on Computer Vision and Pattern Recognition
  (CVPR)}, 2017.

\bibitem{liu2009content}
Feng Liu, Michael Gleicher, Hailin Jin, and Aseem Agarwala,
\newblock ``Content-preserving warps for 3d video stabilization,''
\newblock in {\em ACM Transactions on Graphics}, 2009, vol.~28, p.~44.

\bibitem{lowe2004distinctive}
David~G Lowe,
\newblock ``Distinctive image features from scale-invariant keypoints,''
\newblock {\em International Journal of Computer Vision}, vol. 60, no. 2, pp.
  91--110, 2004.

\bibitem{vedaldi2010vlfeat}
Andrea Vedaldi and Brian Fulkerson,
\newblock ``Vlfeat: An open and portable library of computer vision
  algorithms,''
\newblock {\em http://www.vlfeat.org/}, 2010.

\bibitem{von2010lsd}
Rafael~Grompone Von~Gioi, Jeremie Jakubowicz, Jean-Michel Morel, and Gregory
  Randall,
\newblock ``Lsd: A fast line segment detector with a false detection control,''
\newblock {\em IEEE Transactions on Pattern Analysis and Machine Intelligence},
  vol. 32, no. 4, pp. 722--732, 2010.

\bibitem{fan2012robust}
Bin Fan, Fuchao Wu, and Zhanyi Hu,
\newblock ``Robust line matching through line--point invariants,''
\newblock {\em Pattern Recognition}, vol. 45, no. 2, pp. 794--805, 2012.

\bibitem{fischler1987random}
Martin~A Fischler and Robert~C Bolles,
\newblock ``Random sample consensus: a paradigm for model fitting with
  applications to image analysis and automated cartography,''
\newblock in {\em Readings in computer vision}, pp. 726--740. 1987.

\bibitem{burt1983multiresolution}
Peter~J Burt and Edward~H Adelson,
\newblock ``A multiresolution spline with application to image mosaics,''
\newblock {\em ACM Transactions on Graphics}, vol. 2, no. 4, pp. 217--236,
  1983.

\bibitem{fortun2015optical}
Denis Fortun, Patrick Bouthemy, and Charles Kervrann,
\newblock ``Optical flow modeling and computation: a survey,''
\newblock {\em Computer Vision and Image Understanding}, vol. 134, pp. 1--21,
  2015.

\end{thebibliography}

\end{document}